\documentclass[runningheads]{llncs}

\usepackage{eccv}

\usepackage{eccvabbrv}

\usepackage{graphicx}
\usepackage{booktabs}
\usepackage{makecell}
\usepackage[dvipsnames]{xcolor}

\usepackage[accsupp]{axessibility}  %
\usepackage{epsfig}
\usepackage{graphicx}
\usepackage{amsmath}
\usepackage{amssymb}
\usepackage{verbatim}
\usepackage[dvipsnames]{xcolor}
\usepackage{float}
\usepackage{tabularx}
\usepackage{colortbl}
\usepackage{multirow}
\usepackage{pgfplots}
\usepackage{ragged2e}
\usepackage{overpic}
\usepackage{rotating}
\usepgfplotslibrary{groupplots}

\usepackage[dvipsnames]{xcolor}

\newcommand{\da}[0]{$\downarrow$}
\newcommand{\ua}[0]{$\uparrow$}

\definecolor{FstTable}{HTML}{BDE6CD}
\definecolor{SndTable}{HTML}{E2EEBC}
\definecolor{TrdTable}{HTML}{FFF8C5}

\newcommand{\fst}[1]{\cellcolor{FstTable}#1}
\newcommand{\snd}[1]{\cellcolor{SndTable}#1}
\newcommand{\trd}[1]{\cellcolor{TrdTable}#1}
\newcommand{\cmt}[1]{}

\newcommand{\ucsnetmvs}{UCS-Net \cite{cheng2020deep}}
\newcommand{\patchnetmvs}{PatchMatch-Net \cite{wang2021patchmatchnet}}
\newcommand{\casnetmvs}{CAS-MVSNet \cite{gu2020cascade}}

\newcommand{\ucsnet}{Guided UCS-Net \cite{cheng2020deep}+\cite{poggi2022guided}}
\newcommand{\patchnet}{Guided PatchMatch-Net \cite{wang2021patchmatchnet}+\cite{poggi2022guided}}
\newcommand{\casnet}{Guided CAS-MVSNet \cite{gu2020cascade}+\cite{poggi2022guided}}
\newcommand{\smplrcn}{SimpleRecon \cite{sayed2022simplerecon}}

\newcommand{\nlspn}{NLSPN \cite{nlspn}}
\newcommand{\spagnet}{SpAgNet \cite{conti2023wacv}}
\newcommand{\cmplformer}{CompletionFormer \cite{zhang2023completionformer}}
\newcommand{\ours}{\textbf{DoD (ours)}}

\usepackage{hyperref}

\usepackage{orcidlink}

\begin{document}

\title{Depth on Demand: Streaming Dense Depth from a Low Frame Rate Active Sensor}

\titlerunning{Depth on Demand}

\author{Andrea Conti \orcidlink{0000-0002-0197-0178}$^{1}$ \and
Matteo Poggi \orcidlink{0000-0002-3337-2236}$^{1}$ \and
Valerio Cambareri \orcidlink{0000-0001-6459-6323}$^{2}$ \and \\
Stefano Mattoccia \orcidlink{0000-0002-3681-7704}$^{1}$
}

\authorrunning{A.~Conti et al.}

\institute{
$^{1}$University of Bologna, Italy \\
$^{2}$Sony Depthsensing Solutions, Brussels, Belgium \\
\url{https://andreaconti.github.io/projects/depth_on_demand}}

\maketitle

\begin{figure}
    \centering
    \includegraphics[trim=0cm 12cm 16cm 0cm,clip,width=0.9\textwidth]{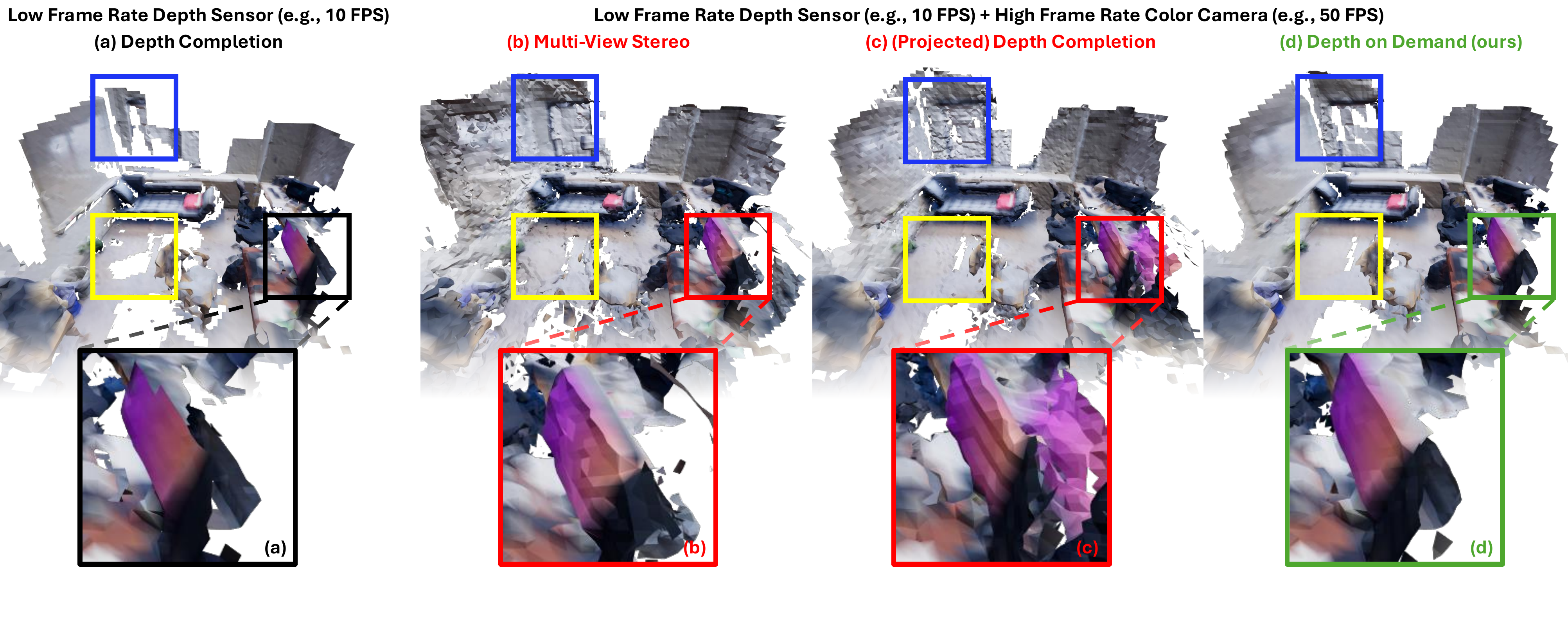}   \captionof{figure}{\textbf{3D reconstruction with a low frame rate, sparse depth sensor.} Running depth completion (a) on low FPS, sparse depth maps generate holes in the final reconstruction. Adding a higher FPS color camera allows for obtaining depth from Multi-View Stereo (b) or projecting depth to nearby color views and running completion (c), with unsatisfactory results. Our framework (d) performs \textit{temporal} completion using two views and one sparse depth frame, yielding denser and more accurate meshes.
   }\label{fig:teaser}
\end{figure}

\begin{abstract}
    High frame rate and accurate depth estimation play an important role in several tasks crucial to robotics and automotive perception. To date, this can be achieved through ToF and LiDAR devices for indoor and outdoor applications, respectively. However, their applicability is limited by low frame rate, energy consumption, and spatial sparsity. Depth on Demand (DoD) allows for accurate temporal and spatial depth densification achieved by exploiting a high frame rate RGB sensor coupled with a potentially lower frame rate and sparse active depth sensor. Our proposal jointly enables lower energy consumption and denser shape reconstruction, by significantly reducing the streaming requirements on the depth sensor thanks to its three core stages: i) multi-modal encoding, ii) iterative multi-modal integration, and iii) depth decoding. 
    We present extended evidence assessing the effectiveness of DoD on indoor and outdoor video datasets, covering both environment scanning and automotive perception use cases.
\end{abstract}    
\section{Introduction}
\label{sec:intro}

We introduce the intrinsic issues related to active depth sensing and how our framework addresses them widening its applicability to different scenarios.

\indent \textbf{Active Depth Sensing.}
In the last decade, RGB-D camera systems have become prominent in fields such as robotics, automotive, and augmented reality, and have scaled down from Kinect v1 to mobile handheld devices such as the Apple iPad. 
In such systems, one or more conventional RGB cameras are coupled with an \textit{active} depth sensor, \ie, a device that leverages active illumination to infer the 3D structure of the framed scene \cite{XIN_IJCV}. 
Among these sensors, Time-of-Flight (ToF) cameras infer the distance by emitting modulated infrared light into the scene and measuring its return time~\cite{XIN_IJCV,Bhandari16, Bamji22}.
On the other hand, Light Detection And Ranging (LiDAR) sensors allow for long-range measurements up to hundreds of meters with or without sunlight at much higher energy consumption and footprint.  

\textbf{Limitations.}
Despite the reconstruction accuracy of active depth sensors, their inherent structure places limits on their usability. ToF sensors are mainly used for mobile devices and can achieve high frame rates, but imply high energy consumption compared to the limited available battery and overheating. Usually, a drastic reduction of their frame rate is required since it corresponds to a drastic reduction in energy consumption. On the other hand, LiDAR sensors are bulky devices mainly used for autonomous driving, and their moving mechanical components (scanning mirrors) limit the frame rate. Finally, both these technologies manifest spatial sparsity, generating meaningful predictions only for specific spatial locations. Such sparsity can intentionally be induced to minimize acquisition time (in scanned LiDAR~\cite{Ma2018SelfSupervisedSS, Bartoccioni23}) or energy consumption (in ToF \cite{Jiang22, Luetzenberg21}). 
Due to these constraints, the adoption of RGB-D camera systems is difficult in various scenarios. Indeed, Augmented Reality (AR) requires extremely low-power camera systems to fit severely constrained energy budgets. Autonomous driving requires high frame rate depth perception to allow reactive safety-critical applications. 3D shape reconstruction from video streams benefits high frame rate reconstruction to achieve dense meshes without requiring very slow movements from the operator.

\textbf{Proposal.}
This paper proposes Depth on Demand (DoD), a framework addressing the three major issues related to active depth sensors in streaming dense depth maps -- \ie spatial sparsity, energy consumption, and limited frame rate. 
The spatial resolution problem represents a well-known issue deeply investigated by the research community through depth completion \cite{survey_completion_2}. On the other hand, energy footprint and low frame rate issues have often been ignored in the literature, although prominent in the deployment of RGB-D systems. We start from the observation that reducing the active sensor temporal resolution -- \ie its frame rate -- power consumption can be modulated accordingly. Indeed, ToF sensors' energy consumption scales almost linearly with frame rate \cite{Chen2018EfficientToF}. DoD allows coupling together an active depth sensor and an RGB camera to stream dense depth at the RGB camera frame rate, which may be much higher than the former one. The benefit is twofold. On the one hand, it allows the adaptation of active depth sensor energy consumption to the task specifications, thus meeting the energy constraints of the ToF use case. On the other, it unlocks frame rates higher than the maximum attainable by the active depth sensor itself -- this effectively tackles the LiDAR use case, where acquisition is limited to 10 Hz while RGB cameras can easily attain 30 Hz or more. Increasing the depth perception frame rate is of great interest in safety-critical applications such as autonomous driving. However, decoupling frame rates benefits also 3D scene reconstruction as it reduces energy consumption and allows for denser reconstructions. This is showcased in Figure~\ref{fig:teaser}: by performing depth completion only at a low frame rate (a) several holes appear in the mesh. Integrating information from a higher frame rate RGB camera (b-d) produces denser meshes. A simple solution to achieve the latter would be relying on Multi-View Stereo algorithms without using depth sensor data (b), or performing depth completion by projecting previous sparse depth points (c). Both these approaches introduce several artifacts as in the highlighted boxes. DoD produces denser and more accurate reconstructions (d).
To summarize, our main contributions are as follows:

\begin{itemize}
    \item We introduce the task of temporal depth stream densification, in which we aim to match the spatial and temporal resolution of a sparse active sensor with one of a higher frame rate RGB camera.
    \item We design a deep architecture devoted to this purpose, exploiting sparse depth measurements and RGB data collected at time $t - n$ to obtain a dense depth map aligned with the RGB frame at time $t$.
    \item We evaluate the proposed framework on several datasets featuring RGB-D video streams and compare it with existing approaches compatible with the outlined setting, proving the superiority of our framework.
\end{itemize}
\section{Related Work}

Our proposal intersects both depth perception from a single RGB-D frame and multiple RGB-only views. Little research exists in the literature concerning the integration of the two for spatiotemporal depth perception densification.

\textbf{Depth Completion.} Depth completion aims at densifying a single monocular sparse depth map obtained by an active depth sensor. These methods can be classified into unguided \cite{sparseinvariantcnns, normalizedcnns, Lu2020FromDW} and RGB-guided techniques \cite{dyspn2022, rignet, penet, Tang_2024_CVPR}.
To date, the state-of-the-art exploits a single RGB-aligned view to better guide the completion procedure and can be distinguished by the propagation method used. Common taxonomies \cite{survey_completion_2,XIN_IJCV} define early-fusion \cite{Dimitrievski2018LearningMO, Imran2019DepthCF, Senushkin2021DecoderMF}, late-fusion \cite{Li2020AMG, rignet, fan2022cascade}, explicit 3D representation \cite{Zhao2020AdaptiveCM}, residual models \cite{Gu2021DenseLiDARAR}, and Spatial Propagation Networks (SPN)-based models \cite{cspn, cspn++, nlspn, dyspn2022, conti2023wacv}.
SPN-based methods are the most effective and have been extensively studied. Initially, \cite{cspn, cspn++} proposed to refine depth through a set of propagating iterations exploiting $3 \times 3$ local affinity matrices learned by a UNet. \cite{nlspn} generalized such a framework introducing deformable sampling in the propagation process. \cite{dyspn2022} introduced an attention-based approach to this sampling, while \cite{yan2024tri} introduced a novel Geometric SPN module. 
On a parallel track, \cite{yan2023desnet} developed an unsupervised framework.

DoD greatly differs from depth completion in that it leverages multi-view cue integration. Indeed, depth completion greatly suffers from input depth outliers, since monocular cues alone are insufficient to effectively filter them out. Conversely, multi-view cues enable ignoring wrong sparse depth points which may occur due to depth projection from a previous depth frame. Moreover, state-of-the-art depth completion methods do not usually integrate techniques to deal with such outliers explicitly \cite{lopez2020project}.

\textbf{Multi-View Stereo.} Multi-view depth perception aims at recovering the structure of scenes given overlapping projections of the 3D space on 2D posed images. It can be applied to reconstructing either a 3D model of an object -- \eg as a point cloud -- or of environments such as indoor scenes. Traditional methods perform such a reconstruction through triangulation and manually engineered features \cite{campbell2008using, furukawa2009accurate, galliani2015massively, schonberger2016pixelwise}. However, state-of-the-art approaches all exploit learned frameworks.
Cost volume-based methods exploit 3D cost volume representation integrated from multiple views. Given a set of depth hypotheses spanning the scene depth range, pixel matching scores are computed over the epipolar lines provided by pose information. Then, 3D convolutional layers are applied as regularization \cite{yao2018mvsnet} to output a depth map \cite{gu2020cascade, yang2020cost, sayed2022simplerecon} aligned with the RGB view.
Volumetric-based methods reconstruct the global scene structure at once back-projecting rays of deep features in a global voxel grid and refining with 3D recurrent layers to finally extract the mesh structure of the scene \cite{sun2021neuralrecon, bozic2021transformerfusion, stier2021vortx, rich20213dvnet, choe2021volumefusion}.

DoD differs from MVS methods not only by architectural design but also in that it integrates both multi-view and depth cues. Multi-view and sparse depth fusion have been scarcely investigated in the literature. \cite{poggi2022guided} injects sparse depth in multi-view stereo networks by modulating their 3D cost volume, following \cite{Poggi_2019_CVPR}. However, such methods struggle to deal with scenes that are not object-centric and usually require a large number of views.
\begin{figure*}[t]
    \centering
    \includegraphics[width=0.30\linewidth]{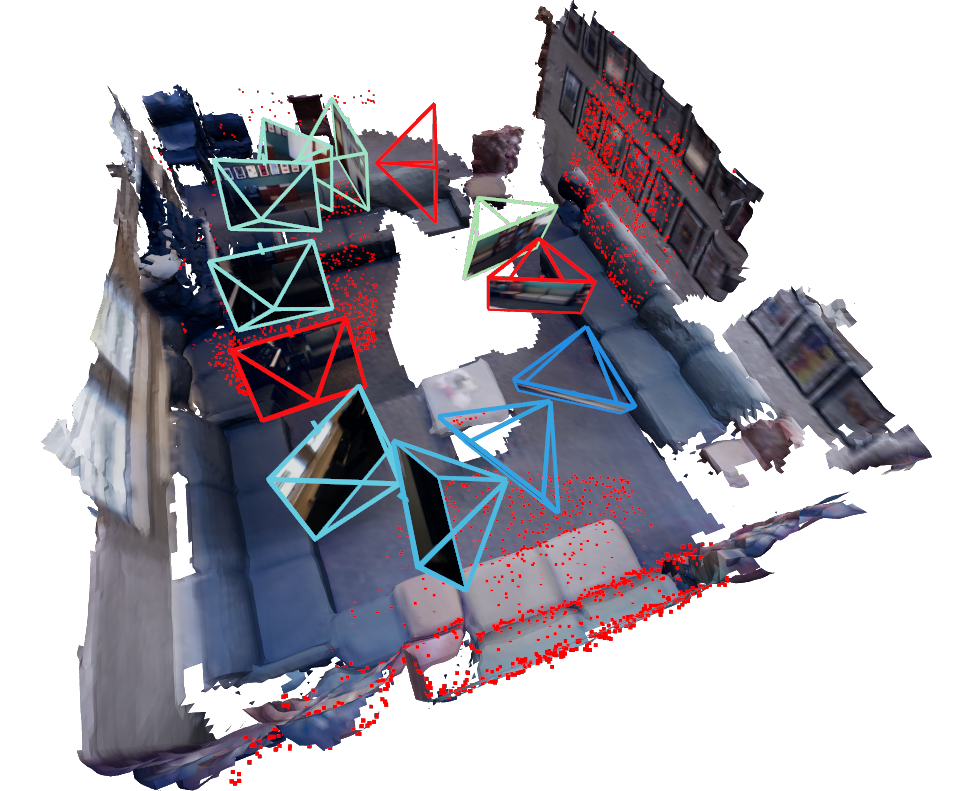}
    \includegraphics[width=0.69\linewidth]{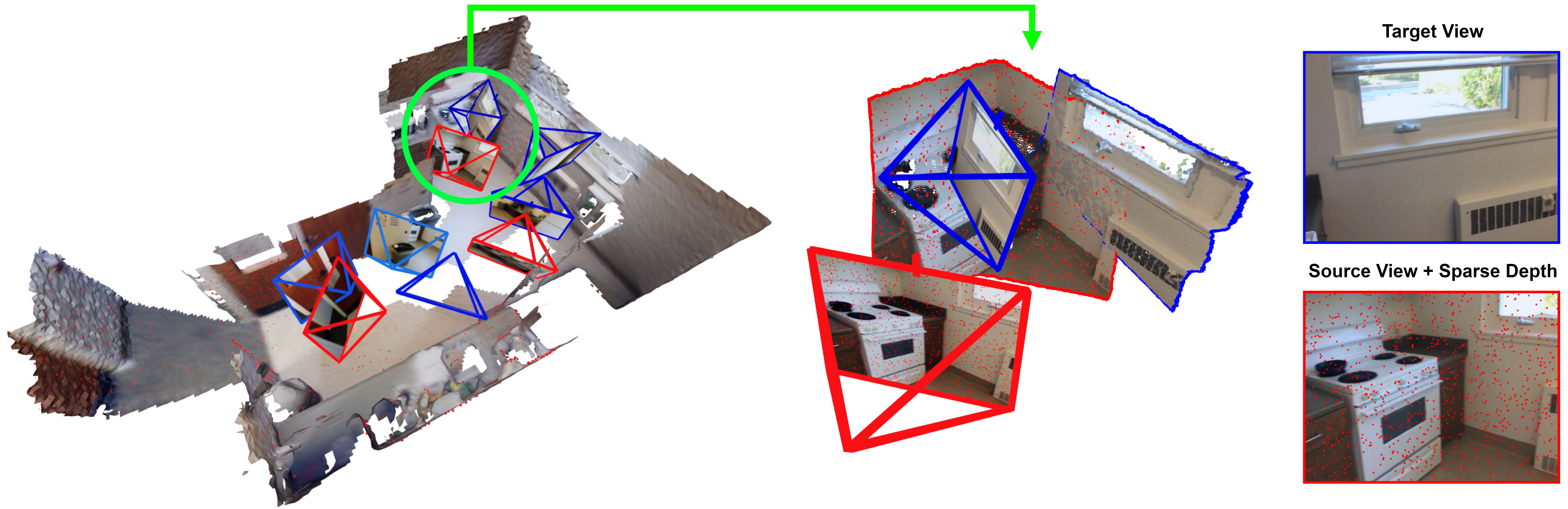}
    \caption{
    \textbf{Temporal Depth Stream Densification Setup.}
    On the left, an example of DoD applied to an indoor video sequence where only a few frames ({\color{red} red} views) are associated with sparse depth data. On the right, a close-up example of the supposed setup. Using an RGB-D video stream with only a few sparse depth frames requires the integration of
    monocular, multi-view, and sparse depth cues. Our framework smoothly enables the recovery of temporal and spatial depth resolution in such a scenario.
    }
    \label{fig:setup-example}
\end{figure*}

\section{Proposed Framework}
\label{sec:proposed framework}

Our approach consists of exploiting the higher framerate of an RGB camera to increase the temporal resolution of an active depth sensor. This is carried out by leveraging multi-view geometry on the RGB video stream and estimating depth for any RGB frame, both those for which measurements from the active sensor are available and those for which are not.
To this aim, we make use of the minimum amount of information needed to exploit geometry -- \ie for each RGB view on which we seek to compute depth (the \textit{target} view) we retain a previously collected RGB frame (\textit{source} view) and sparse depth points (\textit{source} depth). The supposed setup is illustrated in Figure \ref{fig:setup-example}. We conceptually divide our framework into a set of three sequential steps: i) multi-modal encoding, ii) iterative multi-modal integration, and iii) depth decoding. Layer-wise details of the proposed deep modules are provided in the supplementary material.

\subsection{Multi-Modal Encoding}
\label{sec:encoding}

\begin{figure*}[t]
\centering
\includegraphics[width=0.99\linewidth]{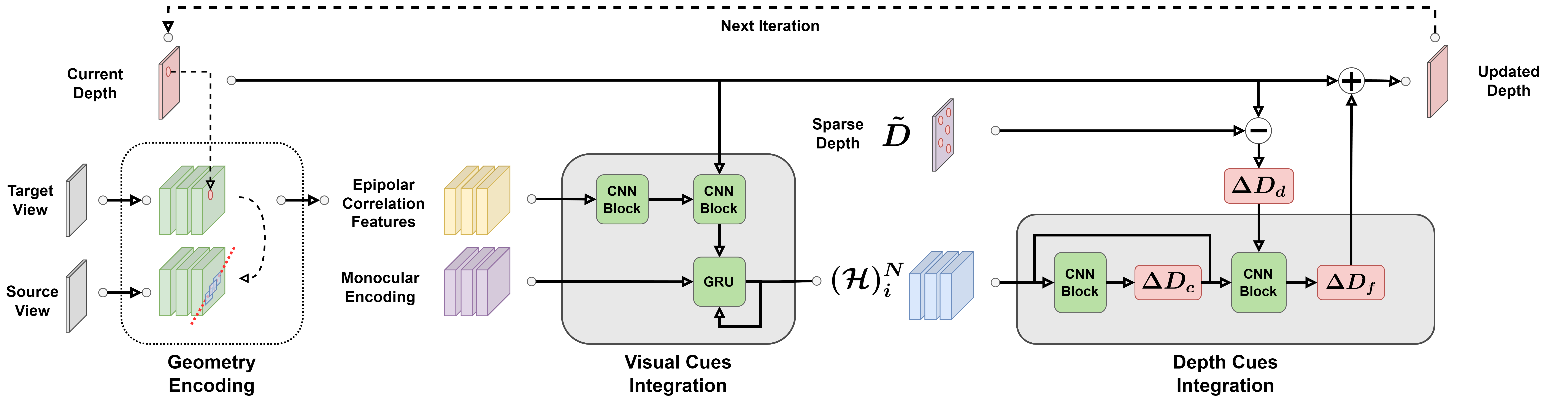}
\caption{\textbf{Depth on Demand Framework Overview.}
We provide a high-level overview of DoD, level-wise architectural details are provided in the supplementary material. DoD embeds multi-view cues and monocular features in the Visual Cues Integration, then integrates sparse depth updates in the Depth Cues Integration. To properly exploit both these information these stages are applied iteratively in the form of depth updates.
}
\label{fig:integration}
\end{figure*}

Our framework exploits information from different \textit{modalities} to perceive 3D structures -- \ie, multi-view geometry, monocular cues, and sparse depth measurements. To this extent, we define our framework as multi-modal. Accordingly, it is important to properly extract useful cues for each of such information sources.
Instead of performing early fusion \cite{gadzicki2020early}, we separately compute domain-appropriate features and delegate a fusion module, detailed in Section \ref{sec:fusion}, to properly integrate them in a common representation. In this section, we specify the encoding of each domain, while Figure \ref{fig:integration} depicts an overall overview of DoD.

\textbf{Geometry Encoding.} 
Multi-view geometry cues stem from the capability to perform matching. We employ the first layers of a ResNet18 \cite{He2015DeepRL} to design a shared encoder, used to extract features $\mathcal{F}^t$,$\mathcal{F}^s$ at $\frac{1}{8}$ spatial resolution, from target and source views respectively. Such features are exploited to compute correlation scores between pixels of the target view and those of the source view. Given the predicted depth at a specific coordinate of the target view $D_{u_t,v_t}$, the matching coordinates of the same point in the source view can be obtained as 

\begin{equation}
    q^s = K P D_{u_t,v_t} K^{-1} q^t
    \label{eq:point-projection}
\end{equation}
where $K$ and $P$ are the camera intrinsic parameters and the relative pose between target and source views, while $q^s = [u_s \ v_s \ 1]^T$ and $q^t = [u_t \ v_t \ 1]^T$ are homogeneous point coordinates in the two frames respectively.
These latter coordinates allow for sampling features from $\mathcal{F}^s$ and $\mathcal{F}^t$ and compute per-point correlation cues as

\begin{equation}
    \mathcal{C} = \frac{1}{\sqrt{F}}\sum_{f=1}^{F}\mathcal{F}^t_{u_tv_tf}\mathcal{F}^s_{u_sv_sf}
    \label{eq:correlation} 
\end{equation}

However, single pixel-wise correlation scores are not sufficient to guide the depth update process effectively. According to multi-view geometry, as the depth value of a pixel in the target image changes, its corresponding matching pixel in the source view is supposed to move along the epipolar line, moving away or approaching the epipole. Meaningful multi-view cues are guaranteed only if the correct updating direction for estimated depth can be inferred. Thus, for each $D_{u_tv_t}$ we sample a set of depth hypotheses relative to the former, moving along the epipolar line. Then, for each of them, we compute a patch of correlation values to increase the distinctiveness of each sampling.
Such procedure generates the ``Epipolar Correlation Features'' represented in the early stages of Figure \ref{fig:integration}. If not otherwise specified, we linearly sample 41 $3 \times 3$ patches within a range of 2 meters, which corresponds to sampling at intervals of 0.1 meters.

\textbf{Monocular Encoding.} Sparse depth information and multi-view correlation data are crucial to deliver accurate 3D reconstructions. Nonetheless, such information fails in the case of moving objects or is not available when large camera pose changes happen, potentially leaving large areas of the field of view empty of information. Thus, it is important to provide a fallback monocular source of information to smoothly complete the not-covered areas.
Purposely, we introduce a monocular encoder exploiting the first layers of a ResNet34 \cite{He2015DeepRL} to output multi-scale feature maps $\tilde{\mathcal{F}}^t_2$, $\tilde{\mathcal{F}}^t_4$ and $\tilde{\mathcal{F}}^t_8$ at respectively $\frac{1}{2}$, $\frac{1}{4}$, $\frac{1}{8}$ resolution out of the target view alone.

\textbf{Sparse Depth Encoding.} Finally, we assume the availability of sparse depth data obtained from an active sensor captured at a previous time instant. We project such sparse depth points onto the target view by means of pose information, obtaining a coarse depth map $\tilde{D}$ that will be exploited for both initialization and iterative multi-modal fusion. Since projecting at a lower resolution may lead to inaccurate positioning when building the sparse depth map, we propagate sub-pixel projection coordinates too. Unlike the depth completion task, projected sparse depth is characterized by errors on moving objects and occlusion, making it more difficult to exploit as a source. We delegate their management to the integration phase, where the exploitation of multiple modalities ameliorates such issues.

\subsection{Multi-Modal Integration}
\label{sec:fusion}

Once features have been extracted for each input modality, we employ a fusion module to combine such information in the common representation of a target-aligned depth map, that is iteratively refined for a fixed number of steps $N$. Our integration module is depicted in Figure \ref{fig:integration} and can be logically divided into two sequential components.

\textbf{Visual Cues Integration.}
The first stage of our fusing module is in charge of extracting depth-related features by visual cues only. Features extracted in the Monocular Encoding step are integrated with the geometric information. This latter consists of a set of correlation features extracted by sampling over the epipolar lines in a relative range with respect to the current depth estimate, as detailed in Section \ref{sec:encoding}. We embed all these cues in the hidden state $(\mathcal{H})_i^N$ of a Gated Recurrent Unit, where $(\cdot)_i^N$ indicates a sequence of tensors across a set of iterations from $i = 0$ to $i = N - 1$. We initialize $(\mathcal{H})_{i=0}^N$ with a deep convolutional module fed with monocular features, specified in the supplementary material.

\textbf{Depth Cues Integration.}
The second stage of our fusing module takes into account sparse depth data availability. First, a branch predicts a depth update $\Delta D_c$ from visual depth-related cues only. Then, a sparse depth update $\Delta D_d$ is computed pixel-wise versus the current prediction as

\begin{equation}
    \Delta D_d = \begin{cases}
    \tilde{D}_{i,j} - D_{i,j} & \text{ if } \tilde{D}_{i,j} > 0 \\
    0                         & \text{ otherwise } \\
    \end{cases}
    \label{eq:delta-update}
\end{equation}
It is worth observing that this latter step generates an update that seeks to refine the current prediction by injecting the exact sparse points, for which zero means either that the depth is correct versus \emph{a priori} information or that the depth measurement is missing for a specific pixel. Finally, the fusion of these updates is carried out by a further branch predicting $\Delta D_f$, which is used to update the current depth prediction. This integration procedure allows for filtering the sparse depth which is likely to contain several outliers due to reprojection -- \eg, as in the case of background points being blended with foreground points at occlusions \cite{aconti2022lidarconf}. Moreover, since the sparse depth data is fused in the update space, missing values can be integrated as zero updates, effectively dealing with the varying sparsity problem often affecting depth completion methods \cite{conti2023wacv}.

\textbf{Iterations and Depth Initialization}. The previously described multi-modal updating strategy is applied multiple times, generating at each iteration a refined depth map that is then used at the subsequent iteration to improve the multi-view correlation samples and the sparse depth update.
Accordingly, an initial depth state is required: we choose to initialize the depth for the first iteration with the sparse depth data, filling the missing coordinates with the mean value of the valid ones. In case no projected sparse depth points are available in the target view we initialize with a reasonable depth value of 3 meters, if not otherwise specified.

\subsection{Depth Decoding}
\label{sec:depth-decoding}

The multi-modal integration module outputs a sequence of incrementally refined depth maps $(D)_i^N$ at $\frac{1}{8}$ resolution. While working at a lower resolution is beneficial in terms of memory and computational time, a method to perform effective upsampling is required.
We exploit a learned procedure inspired by convex upsampling \cite{Teed2020raft}, given the depth map at $\frac{1}{8}$ resolution, we employ a set of three modules $\theta_s(\cdot) \:\: s \in \{2, 4, 8\}$ performing a $2\times$ resolution upsampling composed of two convolutional layers. Each module takes in input the depth map to be upsampled, monocular context information $\tilde{\mathcal{F}}_s \:\: s \in \{2, 4, 8\}$ and a set of features from the previous step, then it outputs $\frac{H}{s} \times \frac{W}{s} \times M$ feature channels and an upsampling mask $\mathcal{W}_s$ of shape $\frac{H}{s}\times\frac{W}{s}\times(2\times2\times9)$. This latter is used to perform a weighted combination over the $3 \times 3$ neighborhood of each depth value normalized by a softmax operation and yields a $2\times$ upsampled depth map. Features are upsampled by nearest neighbor interpolation. The first module $\theta_8(\cdot)$ takes in input the last hidden state $(\mathcal{H})_{i=N-1}^N$. This approach enables both embedding fine-grain monocular contextual information and enforcing locally smooth and consistent depth propagation. Moreover, it allows for a drastic reduction of the upsampling module weights number with respect to conventional convex upsampling. While optimizing, the upsampling module is applied at each iteration for supervision; however, at deploying time it can be used only once for the final prediction, to maximize efficiency.
\pgfplotsset{every tick label/.append style={font=\tiny}}

\section{Experiments}
\label{sec:experiments}

We evaluate our framework in a wide range of scenarios -- \ie indoor video sequences, aerial scenes, automotive environments -- to evaluate its accuracy within single domains, as well as its generalization capabilities. Since each setting manifests its own challenges, we divide the experiments by scenario and highlight the main difficulties faced on a per-dataset basis.

\textbf{Training Protocol.}
To each target frame $I_i$ we associate a buffer of previous $N$ frames $\{(I_j,\ \tilde{D}_j)\ :\ j \in [i-N, i-1]\}$. Then, at each iteration, we randomly select a frame from such buffer as the source one. This is done to augment as much as possible the number of relative poses observed between the source and target view. We apply random color jitter and horizontal flips, adjusting the pose accordingly.
For each sample, we collect a sequence of progressively refined depth maps yielded by the multi-modal integration unit and upsample them to full resolution with the depth decoding approach described in Section \ref{sec:depth-decoding}. We supervise such a sequence of depth maps $(D)_i^N$ using an exponentially decayed $\ell_1$-loss, as described in Equation \ref{eq:loss} with decaying factor $\nu = 0.8$.

\begin{equation}
    L = \sum_{i=1}^N \nu^{N-i} ||(D)_i^N - D_{\text{gt}}||_1
    \label{eq:loss}
\end{equation}

\textbf{Testing Protocol.}
While testing, we suppose an RGB video stream with a higher frame rate than the active depth sensor. Thus, given a sequence of RGB frames $[I_0, \dots, I_n]$ only a few of them will be associated with a depth frame $\{\tilde{D}_0, \dots, \tilde{D}_m\}$. In each testing video sequence, we link to each RGB view $I_i$ the immediately preceding RGB image coupled with a depth frame $(I_j,\ \tilde{D}_j), \, j \le i$ and feed our and competing methods with such data to predict a dense depth map.
By modulating the temporal sparsification ratio between the depth and RGB frames $\tau = f_{\rm D}/f_{\rm RGB}$ we can control how frequently the sparse depth frames are provided; with ratio 1 the task is equivalent to depth completion as depth would be given at every frame.

\textbf{Competitors.}
To fairly compare with existing approaches, we retrain each one following the authors' guidelines but applying the previously described training protocol to increase their robustness to the peculiarities of the proposed task since the standard original training protocol for depth completion struggles at dealing with the considered setup.
Concerning depth completion, we compare with state-of-the-art frameworks \cite{aconti2022lidarconf, nlspn, zhang2023completionformer} projecting sparse depth points from the source frame onto the target view $I_i$. Concerning Multi-View Stereo methods, we select \cite{poggi2022guided} as a natural competitor since it enables standard MVS frameworks \cite{wang2021patchmatchnet, gu2020cascade, cheng2020deep} to exploit both sparse depth and multi-view data natively. Also in this latter case, we train with the aforementioned scheme. Since Multi-View Stereo methods usually exploit a large number of views, we train and evaluate with both 2 and 8 input views.

\begin{table*}[t]
    \centering
    \caption{\textbf{Results on ScanNetV2.} On top, (a) 2D and (b) 3D performance by DoD and competing approaches. At the bottom, (c) 3D performance by DoD with low/high temporal resolution. The \colorbox{FstTable}{best}, \colorbox{SndTable}{second}-best and \colorbox{TrdTable}{third}-best are highlighted.
    }
    \resizebox{\linewidth}{!}{
    \begin{tabular}{c}
    \renewcommand{\tabcolsep}{8pt}
    \begin{tabular}{clcccccccccccc}
    \Xhline{1pt}
    & \multirow{2}{*}{Method}           & \multirow{2}{*}{Views} \cmt{& D         } & \multicolumn{5}{c}{2D Metrics}                                                                      & \multicolumn{6}{c}{3D Metrics}                                                     \\
    \cmidrule(lr){4-8} \cmidrule(lr){9-14}
    & &  & MAE\da      & RMSE\da     & Abs Rel\da  & Sq Rel\da   & $\sigma < 1.05$\ua  \cmt{& $\sigma < 1.25$\ua} & Comp\da     & Acc\da      & Chamfer\da  & Prec\ua     & Recall\ua   & F-Score\ua   \\
    \Xhline{1pt}
    \multirow{4}{*}{\rotatebox{90}{\tiny MVS}} & \smplrcn         & 8  \cmt{&           } & 0.093       & 0.151       & 0.047       & 0.016       & 0.717               \cmt{&                    }& 0.062       & 0.056       & 0.059       & 0.702       & 0.646       & 0.671        \\ 
    & \patchnetmvs & 8 \cmt{&           } & 0.184           & 0.270 & 0.102 & 0.048 & 0.437 & 0.106 & 0.086          & 0.096   & 0.511 &  0.433            & 0.467             \\
    & \casnetmvs   & 8 \cmt{&           } & 0.170           & 0.254 & 0.091 & 0.044 & 0.507 & 0.086 & 0.082     & 0.084 & 0.545 &  0.498         & 0.519             \\
    & \ucsnetmvs   & 8 \cmt{&           }  & 0.167           & 0.252 & 0.088 & 0.042 & 0.512 & 0.084 & 0.082    & 0.083     & 0.547 &  0.502      & 0.522             \\
    \Xhline{1pt}
    \multirow{6}{*}{\rotatebox{90}{\tiny MVS + Depth}} & \patchnet        & 8  \cmt{& \checkmark} & 0.183       & 0.267       & 0.102       & 0.048       & 0.437               \cmt{&                    }& 0.106       & 0.085       & 0.095       & 0.512       & 0.432       & 0.467        \\
    & \casnet          & 8  \cmt{& \checkmark} & 0.124       & 0.203       & 0.068       & 0.029       & 0.635               \cmt{&                    }& 0.064       & 0.061       & 0.062       & 0.667       & 0.634       & 0.649        \\
    & \ucsnet          & 8  \cmt{& \checkmark} & 0.133       & 0.210       & 0.074       & 0.030       & 0.576               \cmt{&                    }& 0.070       & 0.065       & 0.068       & 0.616       & 0.578       & 0.595        \\
    \cmidrule(lr){2-14}
    & \patchnet        & 2  \cmt{& \checkmark} & 0.291       & 0.384       & 0.160       & 0.096       & 0.284               \cmt{& 0.746              }& 0.135       & 0.125       & 0.130       & 0.406       & 0.315       & 0.353        \\
    & \casnet          & 2  \cmt{& \checkmark} & 0.286       & 0.388       & 0.154       & 0.094       & 0.304               \cmt{& 0.752              }& 0.099       & 0.109       & 0.104       & 0.447       & 0.419       & 0.431        \\
    & \ucsnet          & 2  \cmt{& \checkmark} & 0.258       & 0.353       & 0.148       & 0.093       & 0.328               \cmt{& 0.795              }& 0.103       & 0.099       & 0.101       & 0.451       & 0.385       & 0.414        \\
    \Xhline{1pt}
    \multirow{3}{*}{\rotatebox{90}{\tiny Depth}} & \spagnet         & 1  \cmt{& \checkmark} & \trd{0.069} & \trd{0.138} & \trd{0.039} & \snd{0.016} & 0.824               \cmt{& \snd{0.964}        }& \snd{0.046} & \trd{0.037} & \trd{0.042} & 0.836       & 0.789       & 0.810        \\
    & \nlspn           & 1  \cmt{& \checkmark} & \snd{0.067} & \snd{0.137} & \snd{0.037} & \trd{0.017} & \snd{0.847}         \cmt{& \trd{0.963}        }& \snd{0.046} & \snd{0.035} & \snd{0.041} & \snd{0.851} & \snd{0.799} & \snd{0.822}  \\
    & \cmplformer      & 1  \cmt{& \checkmark} & 0.075       & 0.149       & 0.041       & 0.019       & \trd{0.829}         \cmt{& 0.956              }& 0.047       & \trd{0.037} & \trd{0.042} & \trd{0.846} & \trd{0.795} & \trd{0.818}  \\
    \Xhline{1pt}
    & \textbf{DoD (ours)}             & 2  \cmt{& \checkmark} & \fst{0.041} & \fst{0.103} & \fst{0.022} & \fst{0.008} & \fst{0.899}         \cmt{& \fst{0.968}        }& \fst{0.039} & \fst{0.025} & \fst{0.032} & \fst{0.904} & \fst{0.845} & \fst{0.871}  \\
    \Xhline{1pt}
    \multicolumn{3}{c}{}  & \multicolumn{5}{c}{\large{\textbf{(a)}}} & \multicolumn{6}{c}{\large{\textbf{(b)}}} \\
    \end{tabular}
    \\
    \\
    \renewcommand{\tabcolsep}{30pt}
    \begin{tabular}{lcccccc}
    \Xhline{1pt}
    \multirow{2}*{Method} & \multicolumn{6}{c}{3D Metrics} \\
    \cmidrule(lr){2-7}
     & Comp$\downarrow$     & Acc$\downarrow$      & Chamfer$\downarrow$  & Prec$\uparrow$     & Recall$\uparrow$   & F-Score$\uparrow$ \\
    \Xhline{1pt}
    DoD -- Low Temporal Resolution  & \snd{0.064}          & \fst{0.014}          & \snd{0.039}          & \fst{0.961}        & \snd{0.778}        & \snd{0.856} \\
    DoD -- High Temporal Resolution & \fst{0.039}          & \snd{0.025}          & \fst{0.032}          & \snd{0.904}        & \fst{0.845}        & \fst{0.871} \\
    \Xhline{1pt}
    \multicolumn{7}{c}{\large{\textbf{(c)}}} \\
    \end{tabular}
    
    \end{tabular}}
    \label{tab:scannetv2}
\end{table*}
\begin{figure}[t]
    \centering
    \begin{tabular}{c}
    \begin{tabular}{ccccc}
    {\tiny \tiny \textbf Source View} &
    \tiny\tiny{{Target View}} &
    \tiny\tiny{{NLSPN \cite{nlspn}}} &
    \tiny\tiny{{SpAgNet \cite{conti2023wacv}}} &
    \tiny\tiny{{Depth on Demand}} \\
    \includegraphics[width=0.19\linewidth,trim=2cm 2cm 0.3cm 0cm, clip]{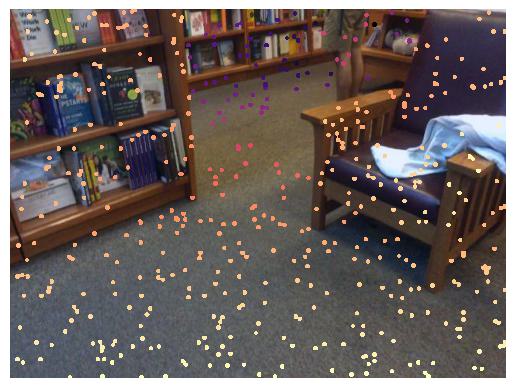}          &
    \includegraphics[width=0.19\linewidth,trim=2cm 2cm 0.3cm 0cm, clip]{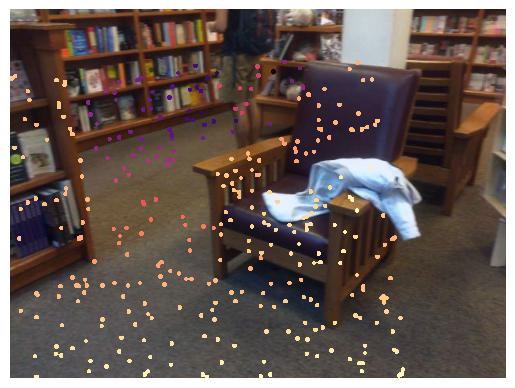}          &
    \includegraphics[width=0.19\linewidth,trim=2cm 2cm 0.3cm 0cm, clip]{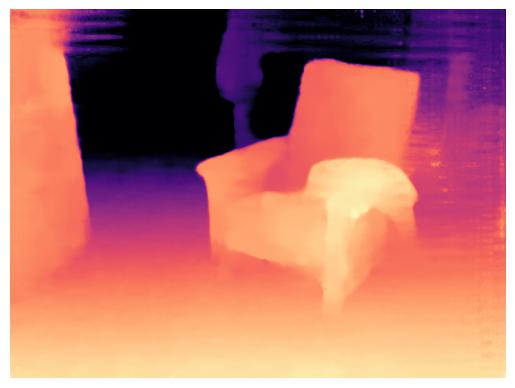}     &
    \includegraphics[width=0.19\linewidth,trim=2cm 2cm 0.3cm 0cm, clip]{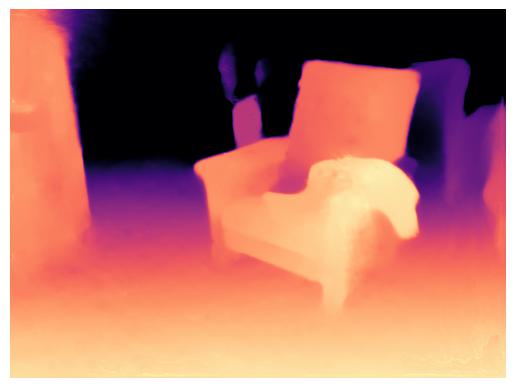}   &
    \includegraphics[width=0.19\linewidth,trim=2cm 2cm 0.3cm 0cm, clip]{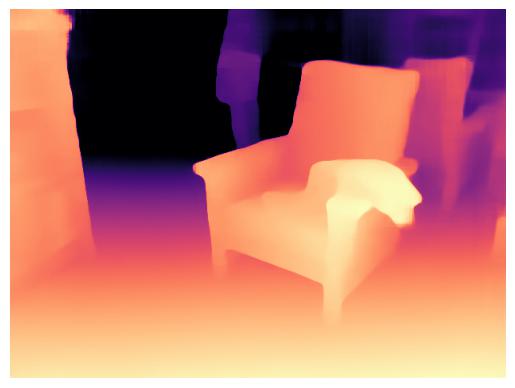}      \\
    \end{tabular}
    \\
    \begin{tabular}{ccc}
    \includegraphics[width=0.33\linewidth, trim=5cm 0cm 5cm 0cm, clip]{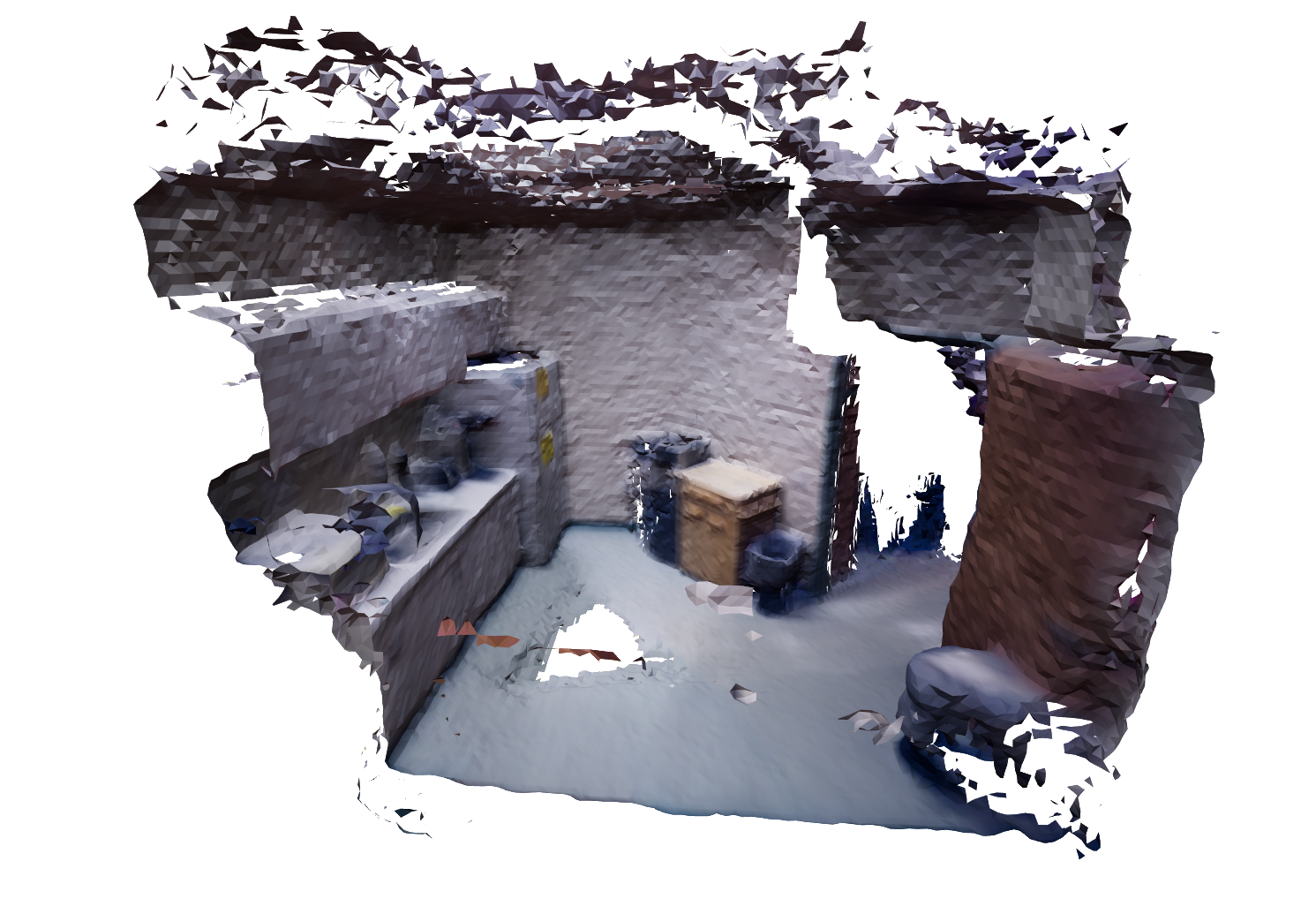}   &
    \includegraphics[width=0.33\linewidth, trim=5cm 0cm 5cm 0cm, clip]{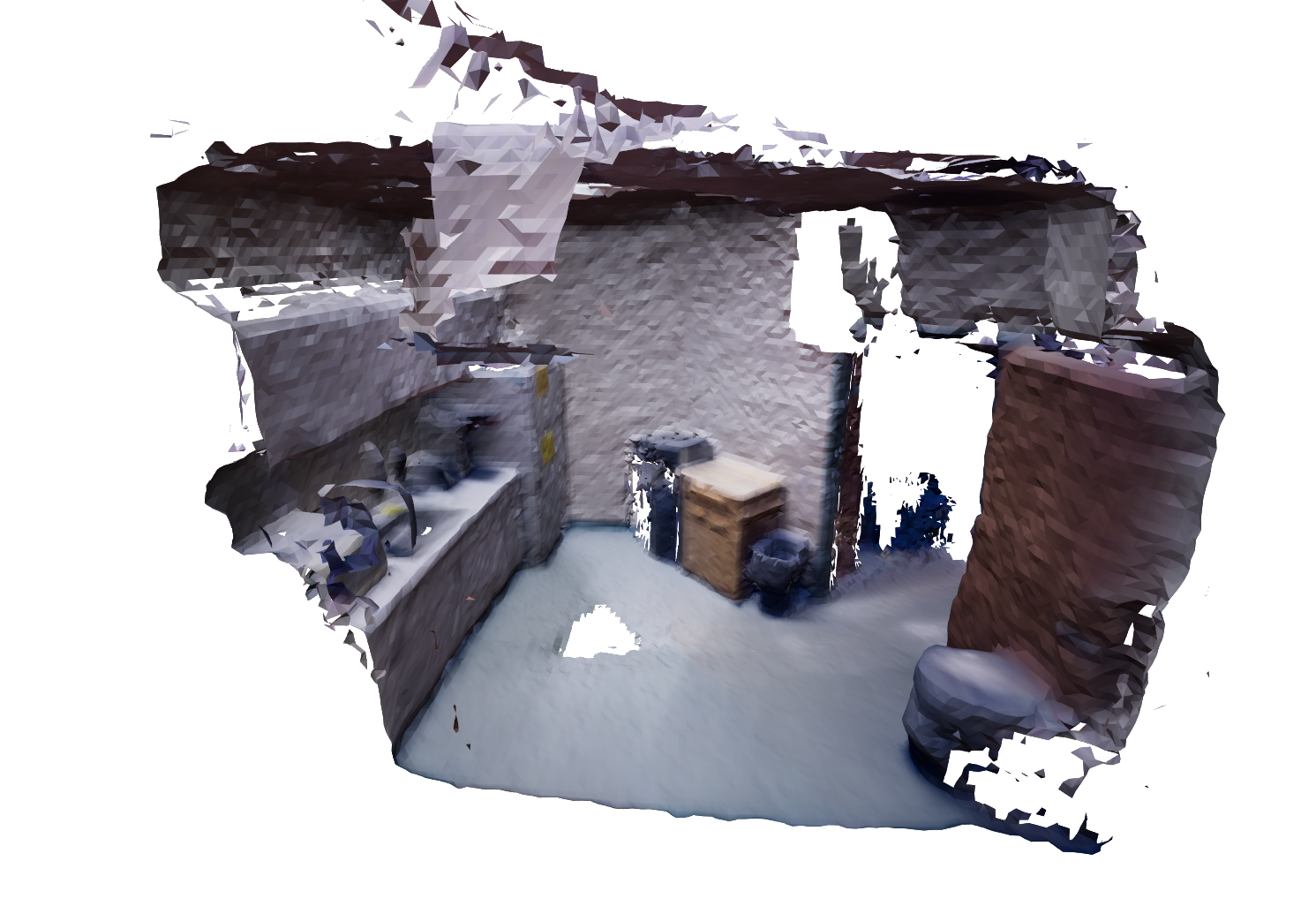} &
    \includegraphics[width=0.33\linewidth, trim=5cm 0cm 5cm 0cm, clip]{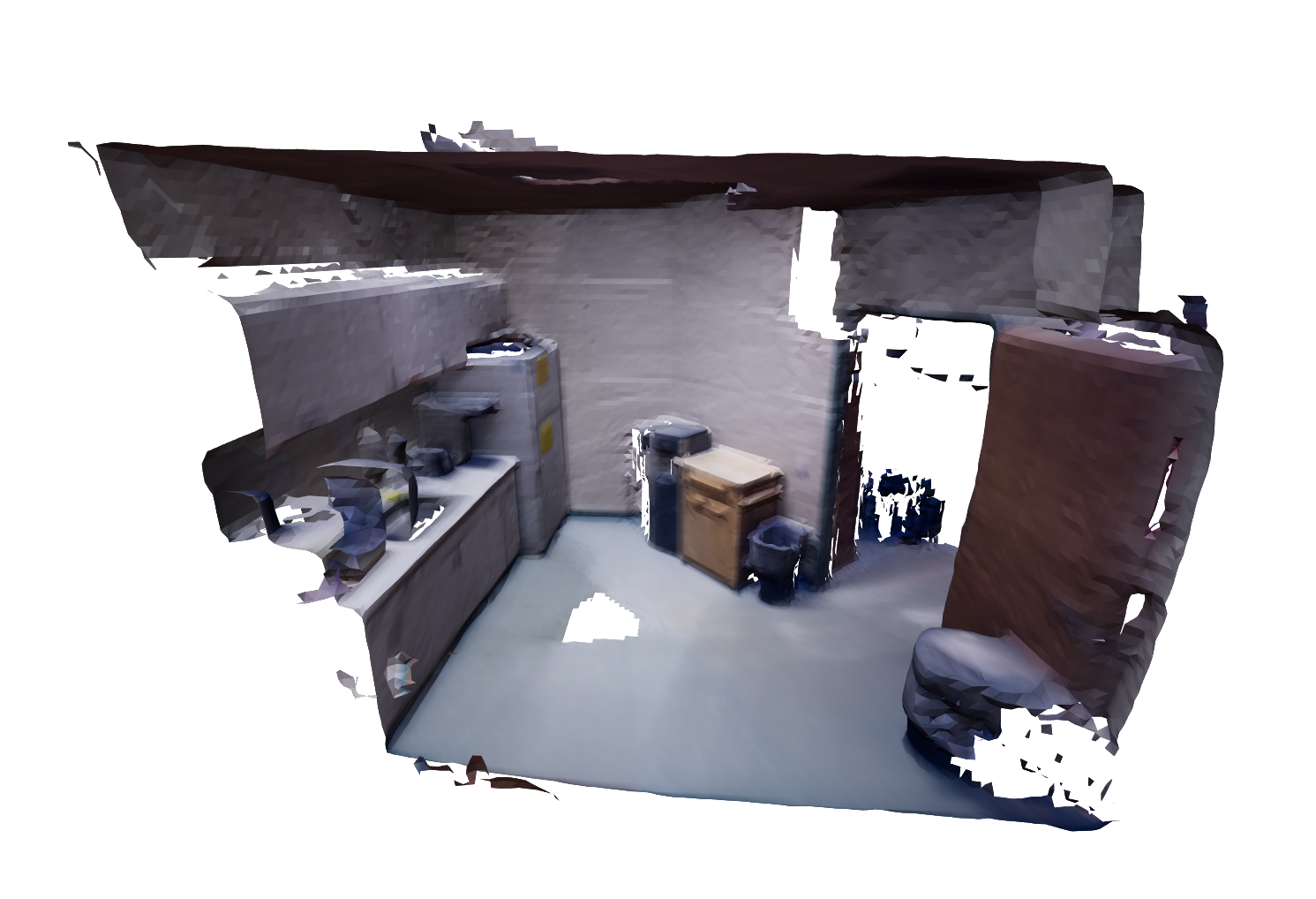}    \\
    \tiny{{NLSPN \cite{nlspn}}} &
    \tiny{{SpAgNet \cite{conti2023wacv}}} &
    \tiny{{Depth on Demand}} \\
    \end{tabular}
    \end{tabular}
    \caption{\textbf{Qualitative results on ScanNetV2.} On top: from left to right the source view with sparse depth points, the target view with projected sparse depth points, and predictions by competitors and DoD. At the bottom: reconstructed meshes by competitors and DoD, 
    respectively at low and high temporal resolution.
    }
    \label{fig:scannetv2-qualitatives}
\end{figure}

\subsection{Indoor Scenario}
For indoor applications, ToF sensors are a popular solution to perceive depth, even though they are characterized by relatively low spatial resolution and a short working range. In this setting, we limit the temporal resolution of the ToF sensor as well, in order to reduce power consumption and overheating. The main challenge in indoor environments is the FoV overlap across consecutive views, which may vary from being almost complete -- \eg, the camera is not moving -- to completely absent.
We train on ScanNetV2 \cite{dai2017ScanNet} and test on both this latter and 7Scenes \cite{glocker2013real-time}, following the protocols described in Section \ref{sec:experiments} by randomly sampling 500 sparse depth points consistently with the depth completion literature \cite{cspn, nlspn, dyspn2022}. For testing, we sparsify depth over time according to $\tau = 0.2$.

\textbf{ScanNetV2.}
ScanNetV2 \cite{dai2017ScanNet} is an RGB-D video dataset containing more than 1500 scans of indoor environments. Table \ref{tab:scannetv2} (a) shows the 2D performance of our framework on standard metrics for depth map evaluation. 
At the top, we report the performance of RGB-only methods \cite{sayed2022simplerecon, wang2021patchmatchnet, gu2020cascade, cheng2020deep} with 8 views in input. Below, we show the performance of \cite{poggi2022guided} with either 8 or 2 input views and projected sparse depth. In such methods, integrating sparse depth in our scenario provides a small improvement, nullified by using only 2 views, \ie the target and a single source view. We ascribe this to their specific design, poor at processing sequential frames. On the contrary, depth completion methods \cite{conti2023wacv, nlspn, zhang2023completionformer} relying only on the target view and projected sparse depth from the source view -- showed at the bottom of Table \ref{tab:scannetv2} -- result in being the most competitive solution for temporal depth stream densification among those existing in the literature.
Eventually, our framework indisputably outperforms completion models on any metric, thanks to the joint use of monocular, multi-view, and sparse depth cues. This superior accuracy of the predicted depth maps also translates into more accurate, dense 3D reconstructions: 
Table \ref{tab:scannetv2} (b) shows how even a few temporally sparse depth measurements largely improve performance versus RGB-only reconstruction carried out by state-of-the-art methods. Again, our framework outperforms existing depth completion solutions by an evident margin. 
Table \ref{tab:scannetv2} (c), instead, highlights the effect of increasing the temporal resolution at which depth is estimated. We can notice how keeping a low temporal resolution -- i.e., the same as the depth sensors -- yields slightly accurate reconstructed meshes, while a higher temporal resolution trades accuracy to increase completeness. Nonetheless, maintaining a high temporal resolution yields better F-Scores overall.
Finally, Figure \ref{fig:scannetv2-qualitatives} shows some qualitative results.

\begin{table}[t]
    \centering
    \caption{\textbf{Results on 7Scenes.} 2D performance by DoD and competing approaches in generalization on 7Scenes. The \colorbox{FstTable}{best}, \colorbox{SndTable}{second}-best and \colorbox{TrdTable}{third}-best are highlighted.}
    \renewcommand{\tabcolsep}{20pt}
    \resizebox{\linewidth}{!}{
    \begin{tabular}{llcccccc}
    \Xhline{1pt}
    & \multirow{2}{*}{Method}           & \multirow{2}{*}{Views} \cmt{& D} &  \multicolumn{5}{c}{2D Metrics} \\
    \cmidrule(lr){4-8}
    & & & MAE\da      & RMSE\da     & Abs Rel\da  & Sq Rel\da   & $\sigma < 1.05$\ua \cmt{& $\sigma < 1.25$\ua }\\
    \Xhline{1pt}

    \multirow{4}{*}{\rotatebox{90}{\tiny MVS}} & \smplrcn         & 8   \cmt{&}            & 0.121      & 0.169       & 0.068       & 0.021       & 0.536              \cmt{& 0.961         }\\
                                               & \patchnetmvs     & 8 \cmt{&           }   & 0.193      & 0.268       & 0.112       & 0.048       & 0.390      \\
                                               & \casnetmvs       & 8 \cmt{&           }   & 0.177      & 0.251       & 0.101       & 0.041       & 0.421      \\
                                               & \ucsnetmvs       & 8 \cmt{&           }   & 0.176      & 0.250       & 0.099       & 0.040       & 0.428      \\
                                               
    \Xhline{1pt}
    \multirow{6}{*}{\rotatebox{90}{\tiny MVS + Depth}} & \patchnet        & 8   \cmt{& \checkmark} & 0.191       & 0.264       & 0.112       & 0.047       & 0.391              \cmt{&               }\\
    & \casnet          & 8   \cmt{& \checkmark} & 0.120       & 0.192       & 0.071       & 0.024       & 0.587              \cmt{&               }\\
    & \ucsnet          & 8   \cmt{& \checkmark} & 0.141       & 0.209       & 0.083       & 0.028       & 0.484              \cmt{&               }\\
    \cmidrule(lr){2-8}
    & \patchnet        & 2   \cmt{& \checkmark} & 0.267       & 0.345       & 0.158       & 0.080       & 0.268              \cmt{&               }\\
    & \casnet          & 2   \cmt{& \checkmark} & 0.250       & 0.338       & 0.141       & 0.069       & 0.303              \cmt{&               }\\
    & \ucsnet          & 2   \cmt{& \checkmark} & 0.228       & 0.306       & 0.139       & 0.063       & 0.303              \cmt{&               }\\
    \Xhline{1pt}
    \multirow{3}{*}{\rotatebox{90}{\tiny Depth}} & \spagnet         & 1   \cmt{& \checkmark} & 0.068       & \trd{0.139} & 0.040       & \snd{0.014} & 0.806              \cmt{&               }\\
    & \nlspn           & 1   \cmt{& \checkmark} & \snd{0.061} & \snd{0.134} & \snd{0.037} & \snd{0.014} & \snd{0.842}        \cmt{&               }\\
    & \cmplformer      & 1   \cmt{& \checkmark} & \trd{0.067} & 0.144       & \trd{0.039} & \trd{0.015} & \trd{0.827}        \cmt{&               }\\
    \Xhline{1pt}
    & \ours            & 2   \cmt{& \checkmark} & \fst{0.043} & \fst{0.106} & \fst{0.025} & \fst{0.008} & \fst{0.896}        \cmt{&               }\\
    \Xhline{1pt}
    \end{tabular}
    }
    \label{tab:7scenes}
\end{table}

\begin{figure}[t]
    \centering
    \begin{tabular}{ccccc}
    \tiny{{Source View}} &
    \tiny{{Target View}} &
    \tiny{{NLSPN \cite{nlspn}}} &
    \tiny{{SpAgNet \cite{conti2023wacv}}} &
    \tiny{{Depth on Demand}} \\
    \includegraphics[width=0.19\linewidth, trim=0.5cm 0.0cm 0.5cm 2.5cm, clip]{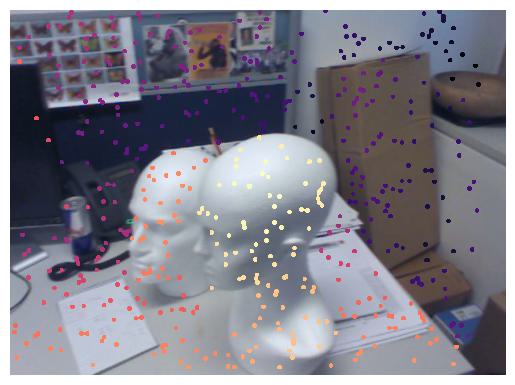}          &
    \includegraphics[width=0.19\linewidth, trim=0.5cm 0.0cm 0.5cm 2.5cm, clip]{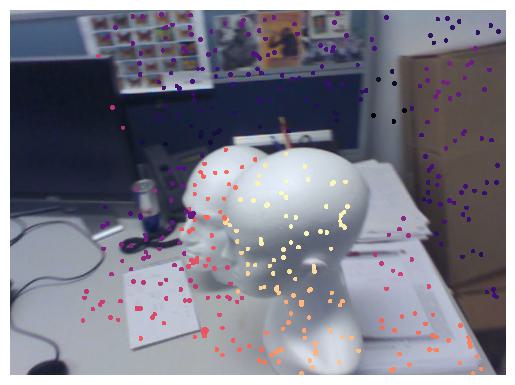}          &
    \includegraphics[width=0.19\linewidth, trim=0.5cm 0.0cm 0.5cm 2.5cm, clip]{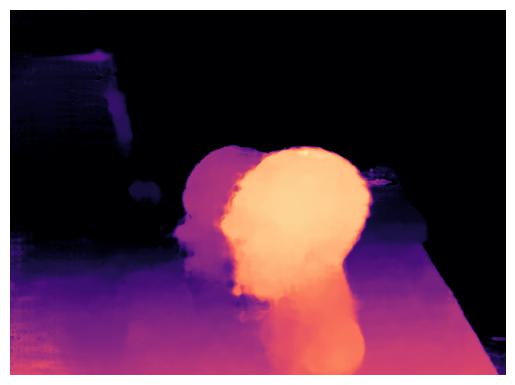}     &
    \includegraphics[width=0.19\linewidth, trim=0.5cm 0.0cm 0.5cm 2.5cm, clip]{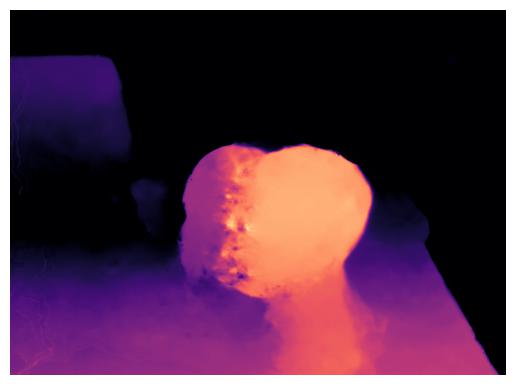}   &
    \includegraphics[width=0.19\linewidth, trim=0.5cm 0.0cm 0.5cm 2.5cm, clip]{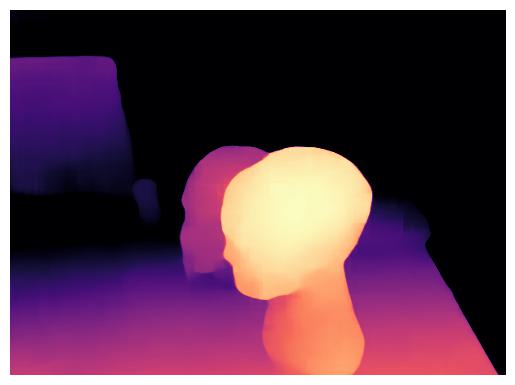}      \\
    \includegraphics[width=0.19\linewidth, trim=2.8cm 2.0cm 4.3cm 3.7cm, clip]{images/qualitatives/heads-seq-02/000013-source.jpg}              &
    \includegraphics[width=0.19\linewidth, trim=2.8cm 2.0cm 4.3cm 3.7cm, clip]{images/qualitatives/heads-seq-02/000013-target.jpg}              &
    \includegraphics[width=0.19\linewidth, trim=2.8cm 2.0cm 4.3cm 3.7cm, clip]{images/qualitatives/heads-seq-02/000013-depth-nlspn.jpg}         &
    \includegraphics[width=0.19\linewidth, trim=2.8cm 2.0cm 4.3cm 3.7cm, clip]{images/qualitatives/heads-seq-02/000013-depth-spagnet.jpg}       &
    \includegraphics[width=0.19\linewidth, trim=2.8cm 2.0cm 4.3cm 3.7cm, clip]{images/qualitatives/heads-seq-02/000013-depth-ours.jpg}          \\
    \end{tabular}
    \caption{\textbf{Qualitative results on 7Scenes.} From left to right: source view with sparse depth points, the target view with projected sparse depth points, and predictions by DoD and existing methods.}
    \label{fig:sevenscenes-qualitatives}
\end{figure}

\textbf{7Scenes.}
We assess the generalization capabilities of our method and the existing alternatives in different indoor environments on the 7Scenes dataset \cite{Shotton2013SceneCR}, by testing the models trained on ScanNetV2 \cite{dai2017ScanNet} without any fine-tuning. Results are collected in Table \ref{tab:7scenes}, where we can notice a trend consistent with what was observed on ScanNetV2: our framework shows remarkable capabilities concerning generalization in the indoor scenario, staying in the lead of the competing approaches. In Figure \ref{fig:sevenscenes-qualitatives} we provide a comparison of handling erroneous sparse depth points due to occlusion where DoD is able to disregard outliers by exploiting multi-view cues.

\subsection{Outdoor Scenario}

3D reconstruction in outdoor environments poses significantly different challenges compared to indoor -- \eg, it features much larger depth ranges and, possibly, scattering of the depth measurements. To study temporal depth completion in this context, we exploit two datasets: TartanAir \cite{tartanair2020iros} and KITTI \cite{Geiger2012CVPR}.

\begin{table}[t]
    \centering
    \caption{\textbf{Results on TartanAir.} 2D performance of our and competing approaches on TartanAir \cite{tartanair2020iros}. The \colorbox{FstTable}{best}, \colorbox{SndTable}{second}-best and \colorbox{TrdTable}{third}-best are highlighted.
    }
    \renewcommand{\tabcolsep}{15pt}
    \resizebox{\linewidth}{!}{
    \begin{tabular}{llcccccc}
    \Xhline{1pt}
    & \multirow{2}{*}{Method}           & \multirow{2}{*}{Views} \cmt{& D} &  \multicolumn{5}{c}{2D Metrics} \\
    \cmidrule(lr){4-8}
    & & & MAE\da      & RMSE\da     & Abs Rel\da  & Sq Rel\da   & $\sigma < 1.05$\ua \cmt{& $\sigma < 1.25$\ua }\\
    \Xhline{1pt}
    \multirow{6}{*}{\rotatebox{90}{\tiny MVS + Depth}} & \patchnet        & 8   \cmt{& \checkmark} & 2.353       & 5.259       & 0.234       & 2.285       & 0.470                 \cmt{&                  }   \\
    & \casnet          & 8   \cmt{& \checkmark} & 1.296       & 3.753       & 0.126       & 1.270       & 0.647                 \cmt{&                  }  \\
    & \ucsnet          & 8   \cmt{& \checkmark} & 1.231       & 3.624       & 0.115       & 1.106       & 0.675                 \cmt{&                  }  \\
    \cmidrule(lr){2-8}
    & \patchnet        & 2   \cmt{& \checkmark} & 3.629       & 6.564       & 0.438       & 3.669       & 0.230                 \cmt{& 0.557            }   \\
    & \casnet          & 2   \cmt{& \checkmark} & 1.985       & 4.845       & 0.185       & 1.794       & 0.492                 \cmt{& 0.803            }  \\
    & \ucsnet          & 2   \cmt{& \checkmark} & 1.804       & 4.513       & 0.177       & 1.526       & 0.486                 \cmt{& 0.812            }  \\
    \Xhline{1pt}
    \multirow{3}{*}{\rotatebox{90}{\tiny Depth}}  & \spagnet         & 1   \cmt{& \checkmark} & \snd{0.841} & \snd{2.273} & \snd{0.090} & \snd{0.561} & \snd{0.718}           \cmt{& \snd{0.914}      }  \\
    & \nlspn           & 1   \cmt{& \checkmark} & \trd{0.941} & \trd{2.327} & \trd{0.113} & \trd{0.623} & \trd{0.613}           \cmt{& \trd{0.904}      }  \\
    & \cmplformer      & 1   \cmt{& \checkmark} & 0.961       & 2.411       & 0.106       & 0.608       & 0.625                 \cmt{& 0.902            }  \\
    \Xhline{1pt}
    & \ours            & 2   \cmt{& \checkmark} & \fst{0.648} & \fst{2.230} & \fst{0.056} & \fst{0.490} & \fst{0.832}           \cmt{& \fst{0.950}      }  \\
    \Xhline{1pt}
    \end{tabular}
  }
    \label{tab:tartanair}
\end{table}

\textbf{TartanAir.}
TartanAir \cite{tartanair2020iros} is a large synthetic dataset featuring photo-realistic environments with different weather and light conditions. It provides a drone-like point of view in a wide set of scenarios featuring high-frequency details and fast camera motion. Table \ref{tab:tartanair} collects the results achieved by existing methods combining multi-view geometry and sparse depth measurements \cite{poggi2022guided} or performing depth completion \cite{nlspn,zhang2023completionformer,conti2023wacv} and our framework. As for the indoor case, completion models largely outperform competitor networks, confirming the limitations of these latter at dealing with the considered problem. Again, our architecture shines in accuracy, achieving the lowest errors by a notable margin.

\begin{figure}[t]
    \centering
    \begin{tabular}{@{}c@{}c@{}c@{}}
    \tiny{{Source View}} & \tiny{{Target View}} & \tiny{{Prediction}} \\
    \includegraphics[width=0.33\linewidth]{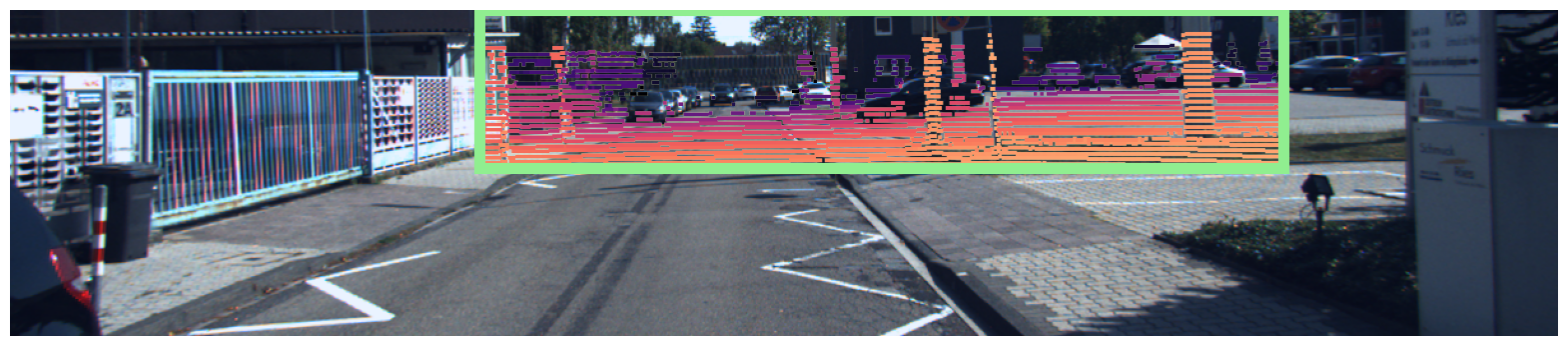}
    & \includegraphics[width=0.33\linewidth]{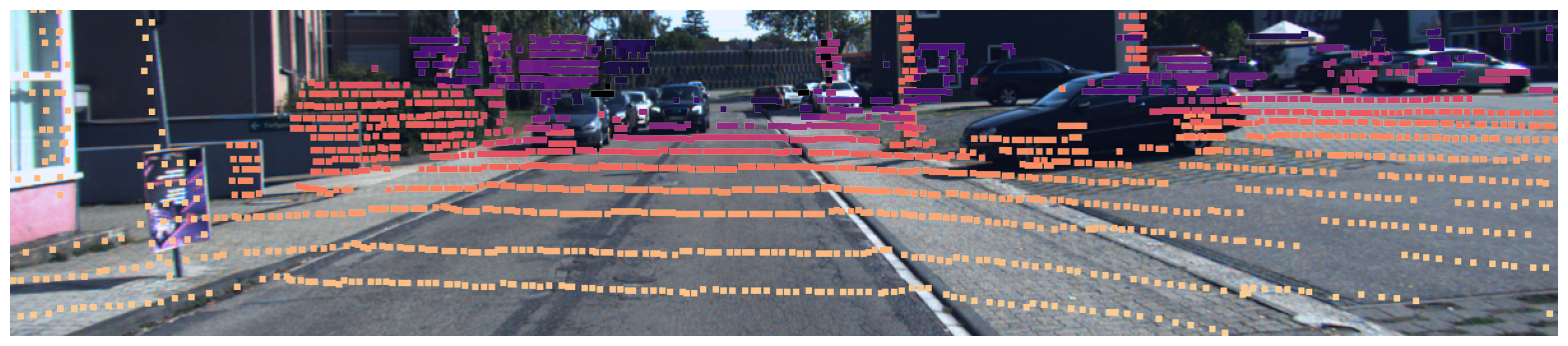}
    & \includegraphics[width=0.33\linewidth]{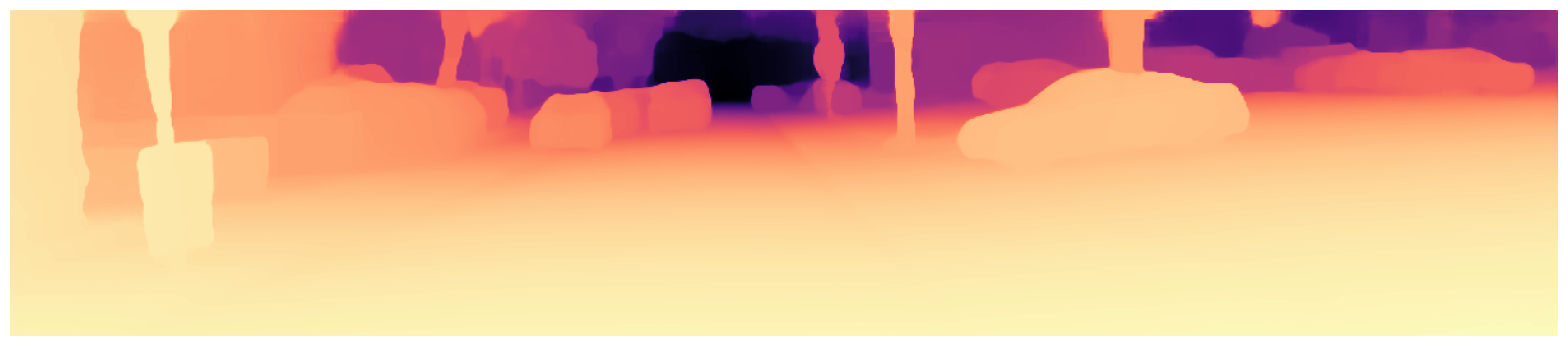}
    \end{tabular}
    \includegraphics[trim=2cm 2cm 2cm 1cm, clip, width=0.9\linewidth]{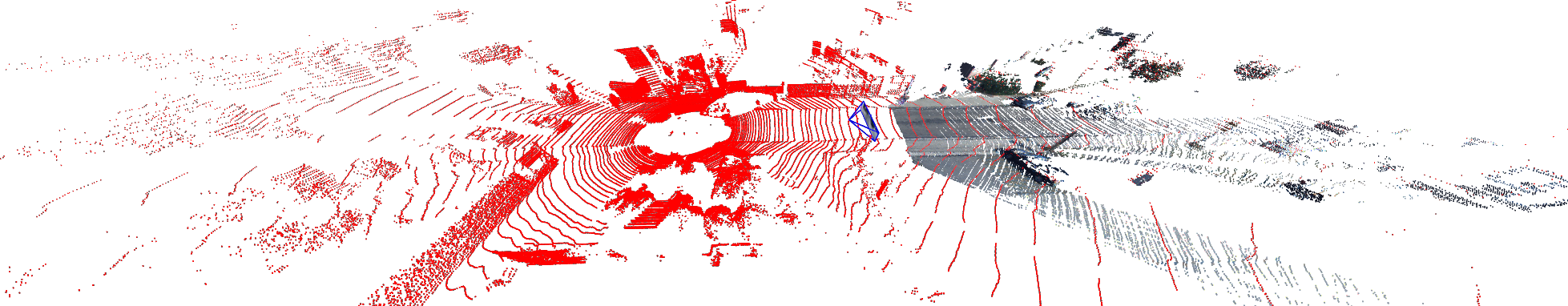}
    \caption{\textbf{KITTI Setup.} On KITTI, we project the 360° LiDAR point cloud over the target point of view. If the camera is moving forward -- as usually happens -- the furthest scan lines are used only, leading to noisy and spaced depth values on the target view. However, the FoV of the target image is usually fully covered.}
    \label{fig:kitti-setup}
\end{figure}

\begin{table}[t]
    \centering
    \caption{\textbf{Results on KITTI.} 2D performance by DoD and competing approaches. The \colorbox{FstTable}{best}, \colorbox{SndTable}{second}-best and \colorbox{TrdTable}{third}-best are highlighted.}
    \renewcommand{\tabcolsep}{8pt}
    \resizebox{\linewidth}{!}{
    \begin{tabular}{ccc}
    \begin{tabular}{llccccccccccc}
    \Xhline{1pt}
    & \multirow{2}{*}{Method}           & \multirow{2}{*}{Views} \cmt{& D} &  \multicolumn{5}{c}{2D Metrics -- \textbf{LiDAR 1Hz}} &  \multicolumn{5}{c}{2D Metrics -- \textbf{LiDAR 0.5Hz}} \\
    \cmidrule(lr){4-8} \cmidrule(lr){9-13}
    & & & MAE\da      & RMSE\da     & Abs Rel\da  & Sq Rel\da   & $\sigma < 1.05$\ua \cmt{& $\sigma < 1.25$\ua } & MAE\da      & RMSE\da     & Abs Rel\da  & Sq Rel\da   & $\sigma < 1.05$\ua \cmt{& $\sigma < 1.25$\ua } \\
    \Xhline{1pt}
    \multirow{6}{*}{\rotatebox{90}{\tiny MVS + Depth}} & \patchnet        & 8  \cmt{& \checkmark} &      2.649  &      5.149  &       0.216 &      3.090  &      0.453         \cmt{&                    } &      2.809  &      5.353  &      0.232  &      3.330  &       0.416        \cmt{&                    }\\
    & \casnet          & 8  \cmt{& \checkmark} &      0.608  &      2.126  &       0.034 &      0.229  &      0.888         \cmt{&                    } &      0.873  &      2.501  &      0.052  &      0.350  &       0.786        \cmt{&                    }\\
    & \ucsnet          & 8  \cmt{& \checkmark} &      0.575  &      1.930  &       0.034 &      0.229  &      0.881         \cmt{&                    } &      0.828  &      2.321  &      0.050  &      0.303  &       0.789        \cmt{&                    }\\
    \cmidrule(lr){2-13}
    & \patchnet        & 2  \cmt{& \checkmark} &      1.898  &      4.165  &       0.117 &      0.863  &      0.496         \cmt{&                    } &      2.282  &      4.564  &      0.145  &      1.154  &       0.404       \cmt{&                    }\\
    & \casnet          & 2  \cmt{& \checkmark} &      0.676  &      2.203  &       0.035 &      0.225  &      0.872         \cmt{&                    } &      0.916  &      2.562  &      0.052  &      0.328  &       0.773       \cmt{&                    }\\
    & \ucsnet          & 2  \cmt{& \checkmark} &      0.545  &      1.859  &       0.030 &      0.146  &      0.885         \cmt{&                    } &      0.837  &      2.330  &      0.049  &      0.277  &       0.779       \cmt{&                    }\\
    \Xhline{1pt}
    \multirow{3}{*}{\rotatebox{90}{\tiny Depth}} & \spagnet         & 1  \cmt{& \checkmark} &      0.532  &      1.626  &       0.027 &      0.095  &      0.879         \cmt{&                    } &      0.687  &      1.865  &      0.037  & \trd{0.133} &       0.808       \cmt{&                    }\\
    & \nlspn           & 1  \cmt{& \checkmark} & \trd{0.426} & \fst{1.282} & \trd{0.023} & \snd{0.069} & \snd{0.902}        \cmt{&                    } & \trd{0.614} & \snd{1.591} & \snd{0.035} & \snd{0.121} & \snd{0.827}       \cmt{&                    }\\
    & \cmplformer      & 1  \cmt{& \checkmark} & \snd{0.348} & \trd{1.299} & \snd{0.019} & \trd{0.085} & \snd{0.939}        \cmt{&                    } & \snd{0.555} & \trd{1.695} & \snd{0.031} & 0.150       & \snd{0.868}       \cmt{&                    }\\
    \Xhline{1pt}
    & \ours            & 2  \cmt{& \checkmark} & \fst{0.347} & \snd{1.288} & \fst{0.017} & \fst{0.061} & \fst{0.944}        \cmt{&                    } & \fst{0.492} & \fst{1.544} & \fst{0.025} & \fst{0.094} & \fst{0.890}       \cmt{&                    }\\
    \Xhline{1pt}
    \end{tabular}
    \end{tabular}
    }
    \label{tab:kitti}
\end{table}

\textbf{KITTI.}
The KITTI \cite{Geiger2012CVPR} dataset is a well-known outdoor benchmark with LiDAR data, widely used for visual odometry, monocular depth prediction, and depth completion.
For automotive applications, 360° LiDAR sensors are usually employed, providing long-range depth at a frequency limited by the revolution time required by the rotating laser beams. Thus, despite the color cameras can acquire frames at a much higher rate, this is usually constrained to the LiDAR frame rate when performing tasks exploiting both -- \eg, depth completion. Nonetheless, when the LiDAR scans are projected from a previous frame over a consequent one as we do to perform temporal depth completion, the target FoV is almost always fully covered with sparse depth points, yet with higher spatial sparsity. Figure~\ref{fig:kitti-setup} shows an example where a LiDAR point cloud collected at a certain time frame is projected over an RGB image collected thereafter, with the camera having moved forward in between the two acquisitions. This causes only the furthest scan lines to be projected over the target view, looking sparser, noisier, and manifesting errors due to occlusions or moving objects. Large areas missing any depth measure may occur in case of occlusion caused by objects in the source view, but still, the FoV is usually fully covered.
In this scenario, our approach is at a disadvantage compared to other methods cause i) the reduced multi-view visual overlap on long distances does not provide large benefits and ii) the spatial distribution of the depth measurements is more steady.

Table \ref{tab:kitti} shows the results achieved by existing methods and ours on this dataset, by simulating different temporal sparsification levels -- \ie RGB camera at $10$Hz and the LiDAR sensor at respectively $1$Hz and $0.5$Hz.
We exploit an off-the-shelf keypoint matcher \cite{lindenberger2023lightglue},
perspective-n-points \cite{Lepetit2009EPnPAA}, and locally-optimized RANSAC \cite{loransac2003} to estimate accurate pose, as already done in the depth completion literature \cite{Ma2018SelfSupervisedSS}. Despite the more challenging setting, our framework still outperforms any existing alternative.

\subsection{Temporal Sparsification Study}

We study the sensitivity of our approach to different \emph{temporal} densities, \ie, frame rate imbalances between the RGB and depth sensor (that is, $\tau = f_{\rm D}/f_{\rm RGB}$). In Figure \ref{fig:memory-time}(a) we report the Mean Absolute Error (MAE) on the 7Scenes test split with the testing protocol described in Section \ref{sec:experiments}, while varying the temporal sparsification $\tau$ from $0.1$ -- \ie one out of ten frames -- to $1$. Actually, when $\tau = 1$ the source view always matches with the target view, and sparse points are aligned with it. Thus, $\tau = 1$ is equivalent to the well-studied depth completion case. This may also occur in real use cases where the camera is static.
We report a sensitivity study to different \emph{spatial}  densities in the supplementary material.

\subsection{Memory and Time Analysis}

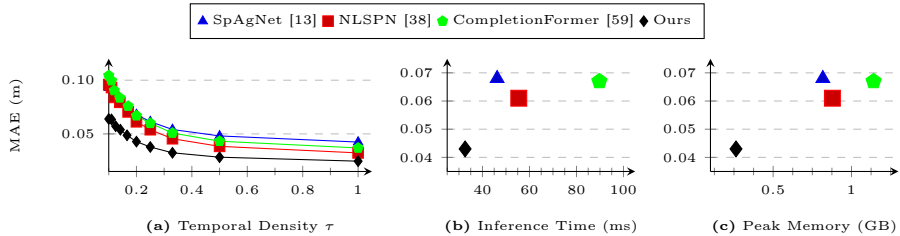
\begin{figure}[t]
    \centering
    \begin{tikzpicture}
    \begin{groupplot}[
        group style={group size=3 by 1},
        axis lines = left,
        legend style = {
            font=\tiny,
            legend columns = 4,
        },
        ymajorgrids=true,
        grid style=dashed,
        scaled ticks = false,
        scale only axis = true,
        yticklabel style={/pgf/number format/.cd, fixed, fixed zerofill, precision=2},
        width=3.0cm,
        height=1.5cm,
    ]

    \nextgroupplot[
        axis lines = left,
        legend style = {
            font=\tiny,
            at={(0.45, 1.05)},
            anchor=south,
        },
        legend columns = 4,
        xlabel = {\tiny \textbf{(a)} Temporal Density $\tau$},
        ylabel = {\tiny MAE (m)},
        ymajorgrids=true,
        grid style=dashed,
        ymin=0.015, ymax=0.12,
        xmax=1.05,
        scaled ticks = false,
        minor x tick num=1,
        height=1.5cm,
        width=3.5cm,
        scale only axis = true,
        yticklabel style={/pgf/number format/.cd, fixed, fixed zerofill, precision=2},
    ]
    \addplot+[mark=triangle*] coordinates {
    (0.10, 0.0930)(0.11, 0.0924)(0.12, 0.0839)(0.14, 0.0803)(0.17, 0.0734)(0.20, 0.0680)(0.25, 0.0614)(0.33, 0.0541)(0.50, 0.0481)(1.00, 0.0424)
    };
    \addplot+[mark=square*] coordinates {
    (0.10, 0.0959)(0.11, 0.0932)(0.12, 0.0840)(0.14, 0.0794)(0.17, 0.0705)(0.20, 0.0613)(0.25, 0.0539)(0.33, 0.0455)(0.50, 0.0384)(1.00, 0.0323)
    };
    \addplot+[green,mark=pentagon*,mark options={fill=green}] coordinates {
    (0.10, 0.1043)(0.11, 0.0996)(0.12, 0.0904)(0.14, 0.0836)(0.17, 0.0759)(0.20, 0.0669)(0.25, 0.0599)(0.33, 0.0508)(0.50, 0.0432)(1.00, 0.0369)
    };
    \addplot+[mark=diamond*] coordinates {
        (0.1,  0.0639)(0.11, 0.0635)(0.125, 0.0572)(0.142, 0.0539)(0.166, 0.0487)(0.2, 0.0428)(0.25, 0.0378)(0.33, 0.0324)(0.5, 0.0283)(1.0, 0.0246)
    };

    \nextgroupplot[
        xlabel = ms,
        ymin=0.035, ymax=0.075,
        height=1.5cm,
        width=2.5cm,
        xmin=25.00, xmax=105,        
        minor x tick num=3,
        mark size=3pt,
        legend to name={CommonLegend},
        xlabel = {\tiny \textbf{(b)} Inference Time (ms)},
    ]
        \addplot+[only marks, mark=triangle*] coordinates {(46.249, 0.068)};
        \addplot+[only marks, mark=square*] coordinates {(55.458, 0.061)};
        \addplot+[green,only marks, mark=pentagon*,mark options={fill=green}] coordinates {(89.740, 0.067)};
        \addplot+[only marks, mark=diamond*] coordinates {(32.625, 0.043)};

    \addlegendimage{}
    \addlegendentry[color=black]{\spagnet\quad}
    \addlegendentry[color=black]{\nlspn\quad}
    \addlegendentry[color=black]{\cmplformer\quad}
    \addlegendentry[color=black]{Ours}

    \nextgroupplot[
        xlabel = {\tiny \textbf{(c)} Peak Memory (GB)},
        ymin=0.035, ymax=0.075,
        width=2.5cm,
        height=1.5cm,
        xmin=0.1, xmax=1.3,     
        minor x tick num=3,
        mark size=3pt,
    ]
        \addplot+[only marks, mark=triangle*] coordinates {(0.817, 0.068)};
        \addplot+[only marks, mark=square*] coordinates {(0.878, 0.061)}; 
        \addplot+[green,only marks, mark=pentagon*,mark options={fill=green}] coordinates {(1.141, 0.067)};
        \addplot+[only marks, mark=diamond*] coordinates {(0.263, 0.043)}; 

    \end{groupplot}
    
    \path (0,0) -- (4.5, 2.0) node{\ref{CommonLegend}};
    
    \end{tikzpicture}
    \caption{\textbf{Memory and Time Study.} We analyze time and memory footprint in evaluation on a single RTX 3090 GPU of our and competing methods 7Scenes.}
    \label{fig:memory-time}
\end{figure}

Figures \ref{fig:memory-time} (b) and \ref{fig:memory-time} (c) report the memory footprint and execution time of the main methods involved in our experiments. We measure the peak memory and computation time the model requires for inference when processing $640 \times 480$ inputs. All measurements are based on a single RTX 3090 GPU and 32-bit floating-point precision. Our approach excels in terms of both.

\subsection{Ablation study}

Finally, we conclude with an ablation study to assess the impact of each module composing DoD. Table \ref{tab:ablation} reports results on ScannetV2 concerning two main studies. In (a), we show how processing multi-view stereo cues and sparse depth impacts the overall accuracy achieved by DoD. Not surprisingly, depth points play a prevalent role, yet alone are insufficient to achieve the best results.
In (b), we report the detailed runtime required by any single component in DoD.

\begin{table}[t]
    \centering
    \caption{\textbf{ Ablation studies.} Experiments on ScanNetV2 aimed at highlighting (a) the impact of multi-view stereo and depth cues, and (b) runtime for each component. }
    \renewcommand{\tabcolsep}{8pt}
    \resizebox{\linewidth}{!}{
    \begin{tabular}{@{}rr@{}}
    \begin{tabular}{cccccc}
        \\
        \\
        \Xhline{1pt}
        \multirow{2}{*}{MVS}& \multirow{2}{*}{Depth} & \multicolumn{2}{c}{2D Metrics} & \multicolumn{2}{c}{3D Metrics} \\
        \cmidrule(lr){3-4} \cmidrule(lr){5-6}
        & & MAE$\downarrow$   & RMSE$\downarrow$ & Chamfer$\downarrow$ & F-Score$\uparrow$ \\
        \Xhline{1pt}
        \checkmark                    &            & \trd 0.187        & \trd 0.257       &  \trd 0.084         &  \trd 0.530       \\
                                      & \checkmark & \snd 0.049        & \snd 0.115       & \snd 0.033          & \snd 0.870        \\
        \checkmark                    & \checkmark & \fst 0.041        & \fst 0.103       & \fst 0.032          & \fst 0.871        \\
        \Xhline{1pt}
        \\
        \\
        \multicolumn{6}{c}{\bf (a)}
        \end{tabular} &
    \begin{tabular}{@{}lccc@{}}
    \Xhline{1pt}
    \multirow{2}{*}{Module} & Single Inf. Time & Calls & Tot. Time \\
    \cmidrule(lr){2-2} \cmidrule(lr){3-3} \cmidrule(lr){4-4}
    & \multicolumn{1}{c}{(ms)} & \multicolumn{1}{c}{(nr.)} & \multicolumn{1}{c}{(ms)} \\
    \Xhline{1pt}
    Geometry Encoding              & 2.060 $\pm$ 0.182            & 1$\times$         & 2.060 $\pm$ 0.182       \\ 
    Monocular Encoding             & 2.175 $\pm$ 0.031            & 1$\times$         & 2.175 $\pm$ 0.031       \\ 
    Correlation Features  & 0.373 $\pm$ 0.001            & 10$\times$        & 3.733 $\pm$ 0.015       \\ 
    Visual Cues Integration        & 1.748 $\pm$ 0.008            & 10$\times$        & 17.480 $\pm$ 0.082       \\
    Depth Cues Integration         & 0.274 $\pm$ 0.005            & 10$\times$        & 2.724 $\pm$ 0.053       \\ 
    Depth Decoding                 & 1.444 $\pm$ 0.040            & 1$\times$         & 1.444 $\pm$ 0.040       \\ 
    \Xhline{1pt}
    Total Time                     &                               &            & 30.196 $\pm$ 0.260       \\ 
    \Xhline{1pt}
    \multicolumn{4}{c}{\bf (b)} \\
    \end{tabular}
    \\    
    \end{tabular}}
    \label{tab:ablation}
\end{table}

\section{Conclusion}

In this paper, we faced the temporal sparsification of a video RGB-D stream when reducing \textit{temporally} the number of depth frames used for accurate 3D reconstruction. This peculiar setup aims to decrease active depth sensors' energy consumption or overcome their limited frame rate compared to cameras. Purposely, we proposed an approach to integrate depth, monocular, and multi-view cues in a remarkably effective common framework, as confirmed by the extensive validation over various datasets. Additionally, our proposal features a significantly lower memory footprint and execution time than competitors.

\textbf{Limitations.} The presence of moving objects inherently harms DoD accuracy, yet without catastrophic failures (see the supplementary material). Future work will focus on this direction to further improve DoD in these occurrences.

{\textbf{Acknowledgment.} We gratefully acknowledge Sony
Depthsensing Solutions SA/NV for funding this research.}

{
    \small
    \bibliographystyle{splncs04}
    \bibliography{egbib}

\begin{thebibliography}{10}
\providecommand{\url}[1]{\texttt{#1}}
\providecommand{\urlprefix}{URL }
\providecommand{\doi}[1]{https://doi.org/#1}

\bibitem{Bamji22}
Bamji, C., Godbaz, J., Oh, M., Mehta, S., Payne, A., Ortiz, S., Nagaraja, S., Perry, T., Thompson, B.: A review of indirect time-of-flight technologies. IEEE Transactions on Electron Devices  \textbf{69}(6),  2779--2793 (2022). \doi{10.1109/TED.2022.3145762}

\bibitem{Bartoccioni23}
Bartoccioni, F., Zablocki, {\'E}., P{\'e}rez, P., Cord, M., Alahari, K.: Lidartouch: Monocular metric depth estimation with a few-beam lidar. Computer Vision and Image Understanding  \textbf{227},  103601 (2023)

\bibitem{Bhandari16}
Bhandari, A., Raskar, R.: Signal processing for time-of-flight imaging sensors: An introduction to inverse problems in computational 3-d imaging. IEEE Signal Processing Magazine  \textbf{33}(5),  45--58 (2016). \doi{10.1109/MSP.2016.2582218}

\bibitem{bozic2021transformerfusion}
Bozic, A., Palafox, P., Thies, J., Dai, A., Nie{\ss}ner, M.: {TransformerFusion}: Monocular {RGB} scene reconstruction using transformers. NeurIPS  (2021)

\bibitem{campbell2008using}
Campbell, N.D., Vogiatzis, G., Hern{\'a}ndez, C., Cipolla, R.: Using multiple hypotheses to improve depth-maps for multi-view stereo. In: European Conference on Computer Vision. pp. 766--779. Springer (2008)

\bibitem{Chen2018EfficientToF}
Chen, Y., Ren, J.S.J., Cheng, X., Qian, K., Gu, J.: Very power efficient neural time-of-flight. 2020 IEEE Winter Conference on Applications of Computer Vision (WACV) pp. 2246--2255 (2018), \url{https://api.semanticscholar.org/CorpusID:56475963}

\bibitem{cheng2020deep}
Cheng, S., Xu, Z., Zhu, S., Li, Z., Li, L.E., Ramamoorthi, R., Su, H.: Deep stereo using adaptive thin volume representation with uncertainty awareness. In: Proceedings of the IEEE/CVF Conference on Computer Vision and Pattern Recognition. pp. 2524--2534 (2020)

\bibitem{cspn}
Cheng, X., Wang, P., Yang, R.: Depth estimation via affinity learned with convolutional spatial propagation network. In: Proceedings of the European Conference on Computer Vision (ECCV). pp. 103--119 (2018)

\bibitem{cspn++}
Cheng, X., Wang, P., Yang, R.: Learning depth with convolutional spatial propagation network. IEEE transactions on pattern analysis and machine intelligence  (2019)

\bibitem{choe2021volumefusion}
Choe, J., Im, S., Rameau, F., Kang, M., Kweon, I.S.: {VolumeFusion}: Deep depth fusion for {3D} scene reconstruction. In: ICCV (2021)

\bibitem{loransac2003}
Chum, O., Matas, J., Kittler, J.: Locally optimized ransac. In: DAGM-Symposium (2003), \url{https://api.semanticscholar.org/CorpusID:15181392}

\bibitem{aconti2022lidarconf}
Conti, A., Poggi, M., Aleotti, F., Mattoccia, S.: Unsupervised confidence for lidar depth maps and applications. In: IEEE/RSJ International Conference on Intelligent Robots and Systems (2022), iROS

\bibitem{conti2023wacv}
Conti, A., Poggi, M., Mattoccia, S.: Sparsity agnostic depth completion. In: Proceedings of the IEEE/CVF Winter Conference on Applications of Computer Vision (WACV). pp. 5871--5880 (January 2023)

\bibitem{dai2017ScanNet}
Dai, A., Chang, A.X., Savva, M., Halber, M., Funkhouser, T., Nie{\ss}ner, M.: Scannet: Richly-annotated 3d reconstructions of indoor scenes. In: Proc. Computer Vision and Pattern Recognition (CVPR), IEEE (2017)

\bibitem{Dimitrievski2018LearningMO}
Dimitrievski, M.D., Veelaert, P., Philips, W.: Learning morphological operators for depth completion. In: Advanced Concepts for Intelligent Vision Systems Conference (2018)

\bibitem{normalizedcnns}
Eldesokey, A., Felsberg, M., Khan, F.S.: Propagating confidences through cnns for sparse data regression. In: British Machine Vision Conference (2018), \url{https://api.semanticscholar.org/CorpusID:44081968}

\bibitem{fan2022cascade}
Fan, R., Li, Z., Poggi, M., Mattoccia, S., et~al.: A cascade dense connection fusion network for depth completion. (2022)

\bibitem{furukawa2009accurate}
Furukawa, Y., Ponce, J.: Accurate, dense, and robust multiview stereopsis. IEEE transactions on pattern analysis and machine intelligence  \textbf{32}(8),  1362--1376 (2009)

\bibitem{gadzicki2020early}
Gadzicki, K., Khamsehashari, R., Zetzsche, C.: Early vs late fusion in multimodal convolutional neural networks. In: 2020 IEEE 23rd international conference on information fusion (FUSION). pp.~1--6. IEEE (2020)

\bibitem{galliani2015massively}
Galliani, S., Lasinger, K., Schindler, K.: Massively parallel multiview stereopsis by surface normal diffusion. In: Proceedings of the IEEE International Conference on Computer Vision. pp. 873--881 (2015)

\bibitem{Geiger2012CVPR}
Geiger, A., Lenz, P., Urtasun, R.: Are we ready for autonomous driving? the kitti vision benchmark suite. In: Conference on Computer Vision and Pattern Recognition (CVPR) (2012)

\bibitem{glocker2013real-time}
Glocker, B., Izadi, S., Shotton, J., Criminisi, A.: Real-time rgb-d camera relocalization. In: International Symposium on Mixed and Augmented Reality (ISMAR). IEEE (October 2013)

\bibitem{Gu2021DenseLiDARAR}
Gu, J., Xiang, Z., Ye, Y., Wang, L.: Denselidar: A real-time pseudo dense depth guided depth completion network. IEEE Robotics and Automation Letters  \textbf{6},  1808--1815 (2021)

\bibitem{gu2020cascade}
Gu, X., Fan, Z., Zhu, S., Dai, Z., Tan, F., Tan, P.: Cascade cost volume for high-resolution multi-view stereo and stereo matching. In: Proceedings of the IEEE/CVF Conference on Computer Vision and Pattern Recognition. pp. 2495--2504 (2020)

\bibitem{He2015DeepRL}
He, K., Zhang, X., Ren, S., Sun, J.: Deep residual learning for image recognition. 2016 IEEE Conference on Computer Vision and Pattern Recognition (CVPR) pp. 770--778 (2015), \url{https://api.semanticscholar.org/CorpusID:206594692}

\bibitem{survey_completion_2}
Hu, J., Bao, C., Ozay, M., Fan, C., Gao, Q., Liu, H., Lam, T.L.: Deep depth completion from extremely sparse data: A survey. IEEE Transactions on Pattern Analysis and Machine Intelligence pp. 1--20 (2022). \doi{10.1109/TPAMI.2022.3229090}

\bibitem{penet}
Hu, M., Wang, S., Li, B., Ning, S., Fan, L., Gong, X.: Towards precise and efficient image guided depth completion  (2021)

\bibitem{Imran2019DepthCF}
Imran, S.M., Long, Y., Liu, X., Morris, D.: Depth coefficients for depth completion. 2019 IEEE/CVF Conference on Computer Vision and Pattern Recognition (CVPR) pp. 12438--12447 (2019)

\bibitem{Jiang22}
Jiang, X., Cambareri, V., Agresti, G., Ugwu, C.I., Simonetto, A., Cardinaux, F., Zanuttigh, P.: A low memory footprint quantized neural network for depth completion of very sparse time-of-flight depth maps. In: Proceedings of the IEEE/CVF Conference on Computer Vision and Pattern Recognition (CVPR) Workshops. pp. 2687--2696 (June 2022)

\bibitem{Lepetit2009EPnPAA}
Lepetit, V., Moreno-Noguer, F., Fua, P.: Epnp: An accurate o(n) solution to the pnp problem. International Journal of Computer Vision  \textbf{81},  155--166 (2009)

\bibitem{Li2020AMG}
Li, A., Yuan, Z., Ling, Y., Chi, W., Zhang, S., Zhang, C.: A multi-scale guided cascade hourglass network for depth completion. 2020 IEEE Winter Conference on Applications of Computer Vision (WACV) pp. 32--40 (2020)

\bibitem{dyspn2022}
Lin, Y., Cheng, T., Zhong, Q., Zhou, W., Yang, H.: Dynamic spatial propagation network for depth completion. vol.~36 (2022). \doi{10.1609/aaai.v36i2.20055}

\bibitem{lindenberger2023lightglue}
Lindenberger, P., Sarlin, P.E., Pollefeys, M.: {LightGlue: Local Feature Matching at Light Speed}. In: ICCV (2023)

\bibitem{lopez2020project}
Lopez-Rodriguez, A., Busam, B., Mikolajczyk, K.: Project to adapt: Domain adaptation for depth completion from noisy and sparse sensor data. In: Proceedings of the Asian Conference on Computer Vision (2020)

\bibitem{Lu2020FromDW}
Lu, K., Barnes, N., Anwar, S., Zheng, L.: From depth what can you see? depth completion via auxiliary image reconstruction. 2020 IEEE/CVF Conference on Computer Vision and Pattern Recognition (CVPR) pp. 11303--11312 (2020), \url{https://api.semanticscholar.org/CorpusID:219615769}

\bibitem{Luetzenberg21}
Luetzenburg, G., Kroon, A., Bj{\o}rk, A.A.: Evaluation of the {Apple} {iPhone} 12 {Pro} {LiDAR} for an {Application in Geosciences}. Scientific Reports  \textbf{11}(1) (Nov 2021). \doi{10.1038/s41598-021-01763-9}, \url{https://doi.org/10.1038/s41598-021-01763-9}

\bibitem{Ma2018SelfSupervisedSS}
Ma, F., Cavalheiro, G.V., Karaman, S.: Self-supervised sparse-to-dense: Self-supervised depth completion from lidar and monocular camera. 2019 International Conference on Robotics and Automation (ICRA) pp. 3288--3295 (2018)

\bibitem{nlspn}
Park, J., Joo, K., Hu, Z., Liu, C.K., Kweon, I.S.: Non-local spatial propagation network for depth completion. In: Proc. of European Conference on Computer Vision (ECCV) (2020)

\bibitem{poggi2022guided}
Poggi, M., Conti, A., Mattoccia, S.: Multi-view guided multi-view stereo. In: IEEE/RSJ International Conference on Intelligent Robots and Systems (2022), iROS

\bibitem{Poggi_2019_CVPR}
Poggi, M., Pallotti, D., Tosi, F., Mattoccia, S.: Guided stereo matching. In: Proceedings of the IEEE/CVF Conference on Computer Vision and Pattern Recognition (CVPR) (June 2019)

\bibitem{XIN_IJCV}
Qiao, X., Poggi, M., Deng, P., Wei, H., Ge, C., Mattoccia, S.: Rgb guided tof imaging system: A survey of deep learning-based methods. Int. J. Comput. Vis.  (2024), \url{https://link.springer.com/article/10.1007/s11263-024-02089-5}

\bibitem{rich20213dvnet}
Rich, A., Stier, N., Sen, P., H{\"o}llerer, T.: 3dvnet: Multi-view depth prediction and volumetric refinement. In: International Conference on 3D Vision (3DV) (2021)

\bibitem{sayed2022simplerecon}
Sayed, M., Gibson, J., Watson, J., Prisacariu, V., Firman, M., Godard, C.: Simplerecon: 3d reconstruction without 3d convolutions. In: Proceedings of the European Conference on Computer Vision (ECCV) (2022)

\bibitem{schonberger2016pixelwise}
Sch{\"o}nberger, J.L., Zheng, E., Frahm, J.M., Pollefeys, M.: Pixelwise view selection for unstructured multi-view stereo. In: European Conference on Computer Vision. pp. 501--518. Springer (2016)

\bibitem{Senushkin2021DecoderMF}
Senushkin, D., Belikov, I., Konushin, A.: Decoder modulation for indoor depth completion. 2021 IEEE/RSJ International Conference on Intelligent Robots and Systems (IROS) pp. 2181--2188 (2021)

\bibitem{Shotton2013SceneCR}
Shotton, J., Glocker, B., Zach, C., Izadi, S., Criminisi, A., Fitzgibbon, A.W.: Scene coordinate regression forests for camera relocalization in rgb-d images. 2013 IEEE Conference on Computer Vision and Pattern Recognition pp. 2930--2937 (2013), \url{https://api.semanticscholar.org/CorpusID:8632684}

\bibitem{stier2021vortx}
Stier, N., Rich, A., Sen, P., H{\"o}llerer, T.: Vortx: Volumetric 3d reconstruction with transformers for voxelwise view selection and fusion. In: International Conference on 3D Vision (3DV) (2021)

\bibitem{sun2021neuralrecon}
Sun, J., Xie, Y., Chen, L., Zhou, X., Bao, H.: {NeuralRecon}: Real-time coherent {3D} reconstruction from monocular video. In: CVPR (2021)

\bibitem{Tang_2024_CVPR}
Tang, J., Tian, F.P., An, B., Li, J., Tan, P.: Bilateral propagation network for depth completion. In: Proceedings of the IEEE/CVF Conference on Computer Vision and Pattern Recognition (CVPR). pp. 9763--9772 (June 2024)

\bibitem{Teed2020raft}
Teed, Z., Deng, J.: Raft: Recurrent all-pairs field transforms for optical flow. In: Vedaldi, A., Bischof, H., Brox, T., Frahm, J.M. (eds.) Computer Vision -- ECCV 2020. pp. 402--419. Springer International Publishing, Cham (2020)

\bibitem{sparseinvariantcnns}
Uhrig, J., Schneider, N., Schneider, L., Franke, U., Brox, T., Geiger, A.: Sparsity invariant cnns. In: 2017 International Conference on 3D Vision (3DV). pp. 11--20 (2017). \doi{10.1109/3DV.2017.00012}

\bibitem{wang2021patchmatchnet}
Wang, F., Galliani, S., Vogel, C., Speciale, P., Pollefeys, M.: Patchmatchnet: Learned multi-view patchmatch stereo. In: Proceedings of the IEEE/CVF Conference on Computer Vision and Pattern Recognition. pp. 14194--14203 (2021)

\bibitem{tartanair2020iros}
Wang, W., Zhu, D., Wang, X., Hu, Y., Qiu, Y., Wang, C., Hu, Y., Kapoor, A., Scherer, S.: Tartanair: A dataset to push the limits of visual slam  (2020)

\bibitem{yan2024tri}
Yan, Z., Lin, Y., Wang, K., Zheng, Y., Wang, Y., Zhang, Z., Li, J., Yang, J.: Tri-perspective view decomposition for geometry-aware depth completion. In: Proceedings of the IEEE/CVF Conference on Computer Vision and Pattern Recognition. pp. 4874--4884 (2024)

\bibitem{rignet}
Yan, Z., Wang, K., Li, X., Zhang, Z., Li, J., Yang, J.: Rignet: Repetitive image guided network for depth completion (2022)

\bibitem{yan2023desnet}
Yan, Z., Wang, K., Li, X., Zhang, Z., Li, J., Yang, J.: Desnet: Decomposed scale-consistent network for unsupervised depth completion. In: Proceedings of the AAAI conference on artificial intelligence. vol.~37, pp. 3109--3117 (2023)

\bibitem{yang2020cost}
Yang, J., Mao, W., Alvarez, J.M., Liu, M.: Cost volume pyramid based depth inference for multi-view stereo. In: Proceedings of the IEEE/CVF Conference on Computer Vision and Pattern Recognition. pp. 4877--4886 (2020)

\bibitem{yao2018mvsnet}
Yao, Y., Luo, Z., Li, S., Fang, T., Quan, L.: Mvsnet: Depth inference for unstructured multi-view stereo. In: Proceedings of the European Conference on Computer Vision (ECCV). pp. 767--783 (2018)

\bibitem{zhang2023completionformer}
Zhang, Y., Guo, X., Poggi, M., Zhu, Z., Huang, G., Mattoccia, S.: Completionformer: Depth completion with convolutions and vision transformers. In: Proceedings of the IEEE/CVF Conference on Computer Vision and Pattern Recognition (CVPR). pp. 18527--18536 (June 2023)

\bibitem{Zhao2020AdaptiveCM}
Zhao, S., Gong, M., Fu, H., Tao, D.: Adaptive context-aware multi-modal network for depth completion. IEEE Transactions on Image Processing  \textbf{30},  5264--5276 (2020)

\end{thebibliography}
}

\clearpage
\setcounter{page}{1}
\maketitlesupplementary

This manuscript provides additional insights about the ECCV paper ``\thetitle''. We collect here additional experimental material, an in-depth description of the modules composing our framework, and qualitative results.

\section{Additional Implementation Details}

\subsection{Architectural Details}

In this section, we deeply detail the core components of our framework -- described in Section 3. Table \ref{tab:architecture-modules} provides implementation details of the main components of our architecture. The visual cues integration module encodes separately the epipolar correlation features and the current depth estimate $(D)_i^N$, then it fuses such data with monocular data $\tilde{\mathcal{F}}^t_8$ using two Gated Recurrent Units with kernel size of $1\times5$ and $5\times1$; this latter choice is done as it leads to a lighter model than using a single $5\times5$ kernel. The depth cues integration module, composed of only four convolutional layers, fuses different depth representations obtained by different sources, as described in Figure 2. The depth decoding module computes multi-scale depth maps. At each iteration a set of upsampling weights and features are computed, then the upsampling is performed using convex upsampling. Finally, the hidden state $(\mathcal{H})_{i=0}^N$ initialization is performed by means of a simple convolutional layer.

\subsection{Training Details}
We train our model on ScanNetV2 \cite{dai2017ScanNet}, TartanAir \cite{tartanair2020iros}, and KITTI \cite{Geiger2012CVPR} with AdamW, $10^{-4}$ learning rate and $10^{-5}$ weight decay. We always perform 100K training steps, dividing the learning rate by 10 at 60K and 90K steps. We train on 2 RTX 3090 in mixed precision with (total) batch size 8. Moreover, we clip gradients with global norm 1 to stabilize the behavior of Gated Recurrent Units and we enforce the same random seed in each training to increase reproducibility. On ScanNetV2 \cite{dai2017ScanNet} we train on the same split defined by \cite{sayed2022simplerecon} with a buffer of 7 source frames to enable consistent comparisons. However, we evaluate on the whole test video sequences subsampled by a factor of ten to mimic a fast-moving camera in an indoor environment; since the camera moves really slow with respect to its high frame rate. We test on 7Scenes \cite{Shotton2013SceneCR} in the same way. On TartanAir \cite{tartanair2020iros} we train and test on the whole video sequences without any frames subsampling using a buffer of 7 source frames while training. Finally, on KITTI \cite{Geiger2012CVPR} we perform training with a buffer of 3 source frames sampled with a frequency of 2Hz.

\begin{table}[t]
    \centering
    \resizebox{\linewidth}{!}{
    \begin{tabular}{cc}
    
        \begin{tabular}{l|llllll}
            \multicolumn{7}{c}{} \\
            \multicolumn{7}{c}{Visual Cues Integration} \\
            \hline
            Input Tensor            & Layer                       & K      & S & In         & Out          & Output Tensor \\
            \hline \hline
            $\mathcal{C}$           & Conv2D + ReLU               & 1      & 1 & $3^2\times41$ & 256 & corr0         \\
            corr0                   & Conv2D + ReLU               & 3      & 1 & 256        & 192           & corr1         \\
            $(D)_i^N$               & Conv2D + ReLU               & 7      & 1 & 1          & 128           & depth0        \\
            depth0                  & Conv2D + ReLU               & 3      & 1 & 128        & 64            & depth1        \\
            depth1,\ corr1          & Conv2D + ReLU               & 3      & 1 & 192+64     & 128-1         & conv0         \\
            conv0, $\tilde{\mathcal{F}}^t_8$, $(\mathcal{H})_i^N$, $(D)_i^N$ & ConvGRU2D                    & (1, 5) & 1 & 128+128+128 & 128           &  hidden0   \\
            hidden0                 & ConvGRU2D                   & (5, 1) & 1 & 128        & 128           & $(\mathcal{H})_{i+1}^N$    \\
            \hline
    
            \multicolumn{7}{c}{} \\
            \multicolumn{7}{c}{Depth Cues Integration} \\
            \hline
            Input Tensor & Layer & K & S & In & Out & Output Tensor \\
            \hline \hline
            $(\mathcal{H})_{i+1}^N$  & Conv2D + ReLU              & 3      & 1 & 128 & 64                    & conv0        \\
            conv0                   & Conv2D                      & 3      & 1 & 64  & 1                      & $\Delta D_c$ \\
            $(\mathcal{H})_{i+1}^N$, $\Delta D_c$, $\Delta D_d$ & Conv2D + ReLU & 3 & 1 & 128+1+1 & 64       & conv1        \\
            conv1                   & Conv2D                      & 3      & 1 & 64 & 1                      & $\Delta D_f$ \\
            \hline
        \end{tabular}
        &
        \begin{tabular}{l|llllll}
            \multicolumn{7}{c}{} \\
            \multicolumn{7}{c}{Depth Decoding} \\
            \hline
            Input Tensor & Layer & K & S & In & Out & Output Tensor \\
            \hline \hline
            $(\mathcal{H})_{i=N-1}^N$, $\tilde{\mathcal{F}}_8^t$, $(D)_{i=N-1}^N$ & Conv2D + ReLU  & 3 & 1 & $128+128+1$     & $3^2\times4+64$    & conv0             \\
            conv0                                                                 & Conv2D         & 3 & 1 & $3^2\times4+64$ & $3^2\times4+64$    & upweights8,feats8 \\
            upweights8, $(D)_{i=N-1}^N$                                           & ConvexUpsample & - & - & $3^2\times4+1$  & 1                  & $D^4$             \\
            $\tilde{\mathcal{F}}_4^t$, $D_4$, feats8                              & Conv2D + ReLU  & 3 & 1 & $64+64+1$       & $3^2\times4+32$    & conv1             \\
            conv1                                                                 & Conv2D         & 3 & 1 & $3^2\times4+32$ & $3^2\times4+32$    & upweights4,feats4 \\
            upweights4, $D_4$                                                     & ConvexUpsample & - & - & $3^2\times4+1$  & 1                  & $D^2$             \\
            $\tilde{\mathcal{F}}_2^t$, $D_2$, feats4                              & Conv2D + ReLU  & 3 & 1 & $64+32+1$       & $3^2\times4$       & conv2             \\
            conv2                                                                 & Conv2D         & 3 & 1 & $3^2\times4$    & $3^2\times4$       & upweights4        \\
            upweights4, $D_2$                                                     & ConvexUpsample & - & - & $3^2\times4$    & 1                  & $D^1$             \\
            \hline
            
            \multicolumn{7}{c}{} \\
            \multicolumn{7}{c}{Hidden State Initialization} \\
            \hline
            Input Tensor               & Layer                       & K      & S & In         & Out          & Output Tensor           \\
            \hline \hline
            $\tilde{\mathcal{F}}_8^t$  & Conv2D + Tanh               & 3      & 1 & 128        & 128          & $(\mathcal{H})_{i=0}^N$ \\
            \\
            \hline
        \end{tabular}
    \end{tabular}
    }
    \vspace{0.2cm}
    \caption{\textbf{Architecture Modules Description.} Description of the main components of our architecture in terms of layers, input and output dimensions. Each line represents a layer of a module where ``Input Tensor'' is the name of the input, ``Layer'' the type of layer used, ``K'' the kernel size if the layer is convolutional, ``S'' the stride if the layer is convolutional, ``In'' the number of input channels, and ``Out'' the number of output channels. Finally, ``Output Tensor'' is the name associated to the output tensor. Name of input and output tensors may refer to intermediate outputs described in the main paper.}
    \label{tab:architecture-modules}
\end{table}

\section{Additional Ablation Studies}

\subsection{Number of Iterations}

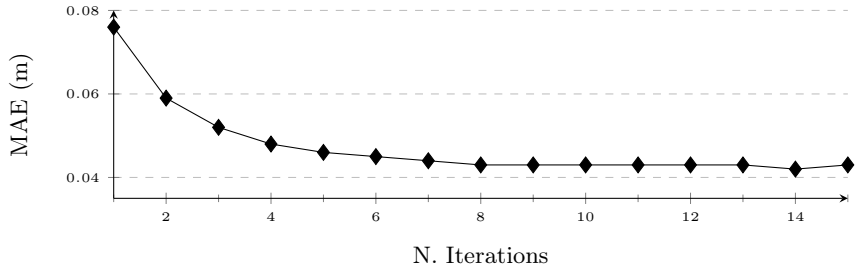
\begin{figure}[h]
\centering
\begin{tikzpicture}
\begin{axis}[
    axis lines = left,
    legend style = {font=\tiny},
    xlabel = {\small N. Iterations},
    ylabel = {\small MAE (m)},
    ymajorgrids=true,
    grid style=dashed,
    ymin=0.035, ymax=0.08,
    scaled ticks = false,
    height=2.5cm,
    mark size=3pt,
    width=0.8\linewidth,
    scale only axis = true,
    yticklabel style={/pgf/number format/.cd, fixed, fixed zerofill, precision=2},
    minor x tick num=1,
]
\addplot[black, mark=diamond*] table[x=n_cycles,y=mae]{data/ncycles_ablation.txt};
\end{axis}
\end{tikzpicture}
\caption{\textbf{Number of Iterations.} Performance using a different number of iterations on 7Scenes \cite{Shotton2013SceneCR}. Our approach allows to change the number of iterations to adapt the time-accuracy trade-off required by the deploying environment.}\vspace{-0.2cm}
\label{fig:iterations-study}
\end{figure}

Our framework is characterized by an iterative module for depth refinement and multi-modal integration, in this section we study the impact of applying a different number of iterations at testing time. During training, we always perform 10 iterations. Figure \ref{fig:iterations-study} shows the mean absolute error in meters on the test split of 7Scenes \cite{Shotton2013SceneCR} performing a different number of iterations. As can be clearly seen the network stabilizes its performance starting from 8 iterations, demonstrating its capability to reach a point of convergence. The number of iterations applied affects the time-accuracy trade-off of our approach. Thus, this latter can be adapted to the deploying requirements modulating such a parameter.

\subsection{Spatial Sparsification}

In this section, we study the performance of our framework applying a variable \textit{spatial} density to the sparse depth data. This is the case in which a different active sensor is used in the deploying environment. Moreover, it allows us to assess the effectiveness of our approach to exploiting multi-view cues. Indeed, when depth data sparsity increases, the unique information other than monocular features our method can leverage to retrieve accurate 3D reconstruction is the information extracted from the RGB source view. When training on ScanNetV2 \cite{dai2017ScanNet}, we always sample 500 sparse depth points. Table \ref{tab:spatial-sparsification-memory-time}(a) reports results obtained with variable density on 7Scenes \cite{Shotton2013SceneCR}, while keeping the temporal density fixed to $\tau = 0.2$. As the spatial density diminishes, depth completion methods struggle since they do not employ geometry cues. On the other hand, our approach is way more robust versus this kind of sparsity since it can recover useful information from the source view as well.

\subsection{Memory and Time}

We provide a detailed memory and time benchmark in Table \ref{tab:spatial-sparsification-memory-time}(b), to complete the overview provided in Figure 7 in the main paper, where only the methods providing the best trade-off between memory, time, and accuracy are highlighted. Notably, PatchMatchNet~\cite{wang2021patchmatchnet} using only 2 views is lighter and faster than our approach. However, with respect to our approach, it provides disastrous performance since the MAE error is $6\times$ worse, with comparable memory occupancy and a small $1.7\times$ speed boost.

\begin{table}[t]
    \centering
    \renewcommand{\tabcolsep}{8pt}
    \resizebox{\linewidth}{!}{
    \begin{tabular}{@{}cc@{}}

    \begin{tabular}{@{}c@{}}

    \begin{tabular}{l|ccccc}
    \multicolumn{6}{c}{200 Points} \\
    \hline
    Method &          MAE\da &         RMSE\da &      Abs Rel\da &       Sq Rel\da & $\sigma < 1.05$\ua \\
    \hline \hline
       \spagnet &     \trd  0.079 &     \snd  0.150 &      \trd 0.048 &     \snd  0.016 &           0.760 \\
         \nlspn &     \snd  0.075 &     \trd  0.152 &      \snd 0.046 &     \trd  0.017 &      \trd 0.794 \\
    \cmplformer &           0.081 &           0.161 &      \trd 0.048 &           0.018 &      \snd 0.777 \\
          \ours &     \fst 0.050  &     \fst  0.117 &      \fst 0.029 &     \fst 0.010  &      \fst 0.872 \\
    \hline
    \end{tabular}

    \\ \\

    \begin{tabular}{l|ccccc}
    \multicolumn{6}{c}{100 Points} \\
    \hline
         Method &          MAE\da &         RMSE\da &      Abs Rel\da &       Sq Rel\da & $\sigma < 1.05$\ua \\
    \hline \hline
       \spagnet &    \trd   0.094 &     \snd  0.165 &     \snd  0.056 &    \snd   0.018 &           0.696 \\
         \nlspn &    \snd   0.093 &     \trd  0.171 &     \snd  0.056 &    \trd   0.021 &     \snd  0.725 \\
    \cmplformer &           0.099 &           0.181 &           0.059 &           0.023 &     \trd  0.703 \\
          \ours &    \fst   0.061 &     \fst  0.129 &     \fst  0.035 &    \fst   0.011 &     \fst  0.832 \\
    \hline
    \end{tabular}
    
    \\ \\

    \begin{tabular}{l|ccccc}
    \multicolumn{6}{c}{50 Points} \\
    \hline
         Method &          MAE\da &         RMSE\da &      Abs Rel\da &       Sq Rel\da & $\sigma < 1.05$\ua \\
    \hline \hline
       \spagnet &    \snd   0.116 &    \snd   0.187 &     \snd  0.070 &    \snd   0.023 &     \trd  0.600 \\
         \nlspn &    \trd   0.118 &    \trd   0.198 &     \trd  0.071 &    \trd   0.026 &     \snd  0.616 \\
    \cmplformer &           0.127 &           0.211 &           0.075 &           0.029 &           0.594 \\
          \ours &    \fst   0.080 &    \fst   0.152 &     \fst  0.045 &    \fst   0.015 &     \fst  0.757 \\
    \hline
    \end{tabular}
    
    \end{tabular}

    &
    
    \begin{tabular}{lc|ccc}
    \\ \\ \\ \\ \\ \\ \\ \\ \\
    \hline
    Method            & Nr.           & MAE\da       & Time (ms)           & Memory (GB)            \\
    \hline \hline
    \patchnet         & 8             &  0.191       &  51.542             &  \trd 0.321            \\
    \casnet           & 8             &  0.120       &  95.556             &  1.100                 \\
    \ucsnet           & 8             &  0.141       &  104.63             &  1.064                 \\
    \patchnet         & 2             &  0.267       &  \fst 19.362        &  \fst 0.224            \\
    \casnet           & 2             &  0.250       &  50.750             &  0.983                 \\
    \ucsnet           & 2             &  0.228       &  55.668             &  1.063                 \\
    \spagnet          & 1             &  0.068       &  \trd 46.249        &  0.817                 \\
    \nlspn            & 1             &  \snd 0.061  &  55.458             &  0.878                 \\
    \cmplformer       & 1             &  \trd 0.067  &  89.741             &  1.141                 \\
    \ours             & 2             &  \fst 0.043  &  \snd 32.625        &  \snd 0.263            \\ 
    \hline
    \end{tabular}

    \\ \\ {\large \bf (a)} & {\large \bf (b)}

    \end{tabular}

    }
    \caption{\textbf{Spatial Sparsification and Memory-Time Studies.} On the left, Spatial sparsification study on 7Scenes \cite{Shotton2013SceneCR}. We keep fixed the sparsification ratio $\tau = 0.2$ and progressively reduce the number of sparse depth points projected. Our approach is the most robust versus spatial sparsification since it enables multi-view cues exploitation.
    On the right, we study the memory and time impact of our approach. DoD provides the best trade-off performance, leading the accuracy ranking by a large margin and still being extremely lightweight in terms of memory and inference time.}
    \label{tab:spatial-sparsification-memory-time}
\end{table}

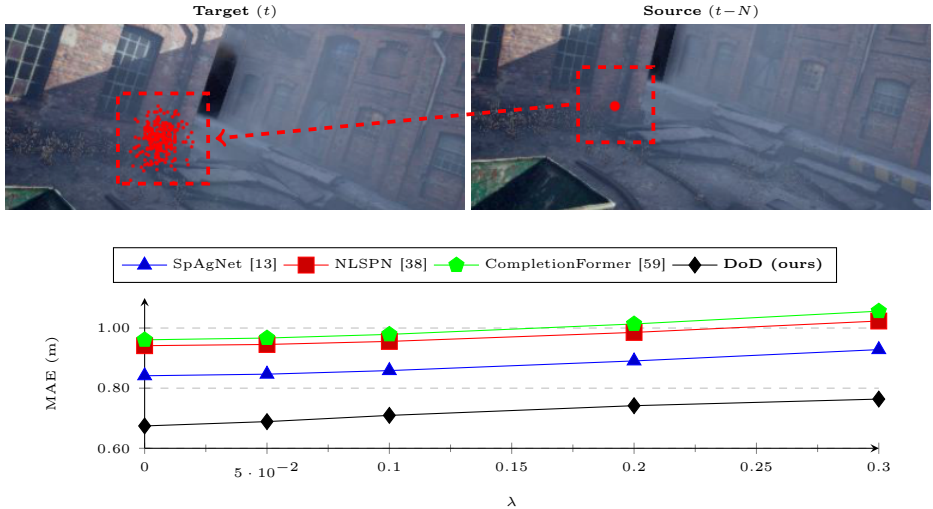
\begin{figure}
\centering
\begin{tabular}{@{}c@{}}
\begin{tabular}{@{}cc@{}}
    \tiny \textbf{Target} ($\scriptstyle t$) & \tiny \textbf{Source} ($\scriptstyle t - N$) \\
    
    \begin{tikzpicture}
        \node[anchor=south west,inner sep=0] at (0,0) {\includegraphics[width=0.50\linewidth,trim={2cm 3.5cm 4cm 4.5cm},clip]{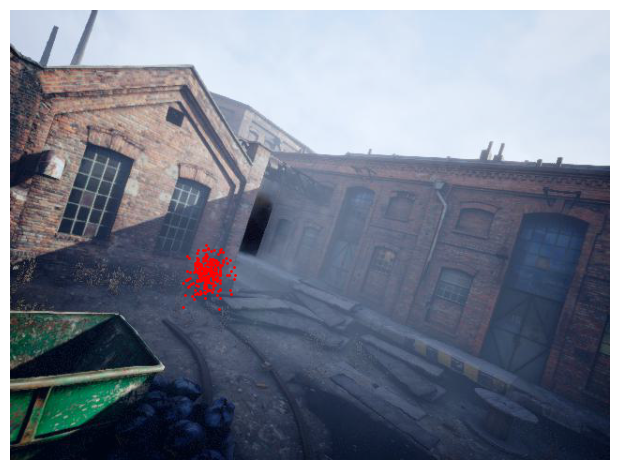}};
        \draw[red, line width=0.05cm, dashed] (1.5, 0.35) rectangle ++(1.2, 1.2);
    \end{tikzpicture}
    &
    \begin{tikzpicture}
        \node[anchor=south west,inner sep=0] at (0,0) {\includegraphics[width=0.50\linewidth,trim={2cm 3.5cm 4cm 4.5cm},clip]{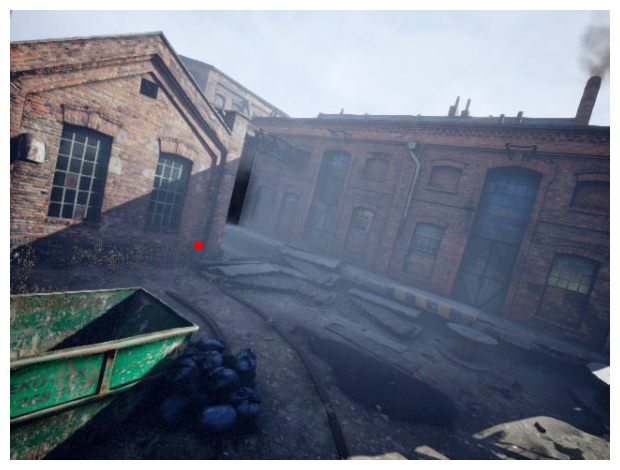}};
        \draw[red, line width=0.05cm, dashed] (1.42, 0.9) rectangle ++(1.00, 1.00);
        \begin{scope}[overlay]
            \draw [red, line width=0.05cm, dashed, ->] (1.3, 1.4) -- (-3.4, 0.95);
        \end{scope}
    \end{tikzpicture}
\end{tabular}
\\
\\
\begin{tikzpicture}
\begin{axis}[
    axis lines = left,
    legend style = {
        font=\tiny \tiny,
        at={(0.45, 1.10)},
        anchor=south,
    },
    legend columns = 4,
    xlabel = {\tiny $\lambda$},
    ylabel = {\tiny MAE (m)},
    ymajorgrids=true,
    grid style=dashed,
    scaled ticks = false,
    height=2.0cm,
    mark size=3pt,
    width=0.8\linewidth,
    ymin=0.6,
    ymax=1.1,
    scale only axis = true,
    yticklabel style={/pgf/number format/.cd, fixed, fixed zerofill, precision=2},
    minor x tick num=1,
]
\addplot+[mark=triangle*] table[x=noise,y=spagnet]{data/pose_noise.txt};
\addlegendentry{\spagnet}
\addplot+[mark=square*] table[x=noise,y=nlspn]{data/pose_noise.txt};
\addlegendentry{\nlspn}
\addplot+[green,mark=pentagon*,mark options={fill=green}] table[x=noise,y=cmplformer]{data/pose_noise.txt};
\addlegendentry{\cmplformer}
\addplot+[mark=diamond*] table[x=noise,y=ours]{data/pose_noise.txt};
\addlegendentry{\ours}
\end{axis}
\end{tikzpicture}
\end{tabular}
\vspace{-0.5cm}
\caption{\textbf{Pose Noise Sensitivity Study.} We assess the impact of noisy pose in our and competitor frameworks perturbing pose information with Gaussian noise. At the top, we show qualitatively how such a noise affects depth point projection. At the bottom, we evaluate methods performance versus noise intensity on TartanAir \cite{tartanair2020iros}. Each model tested is trained with noise-free pose.}
\label{fig:pose-noise-study}
\end{figure}

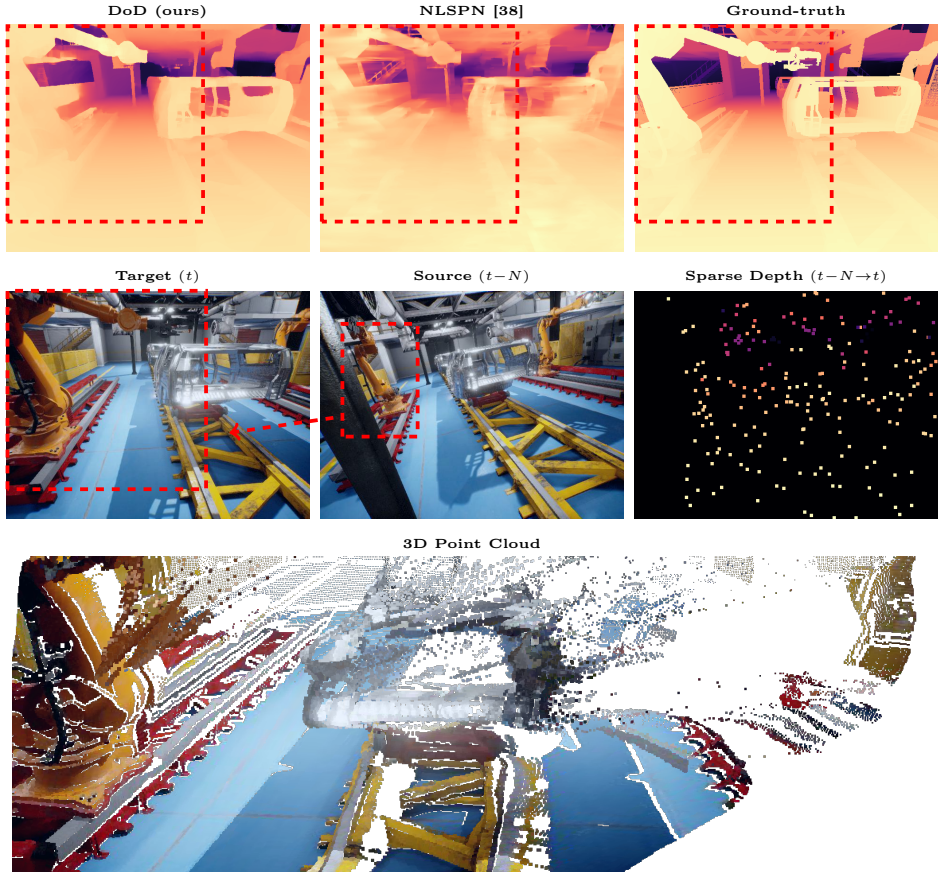
\begin{figure}[t]
    \centering
    \renewcommand{\tabcolsep}{2pt}
    \begin{tabular}{@{}ccc@{}}
    \tiny \textbf{DoD (ours)} & \tiny \textbf{\nlspn} & \tiny \textbf{Ground-truth}                                         \\
    \begin{tikzpicture}
        \node[anchor=south west,inner sep=0] at (0,0) {\includegraphics[width=0.33\linewidth]{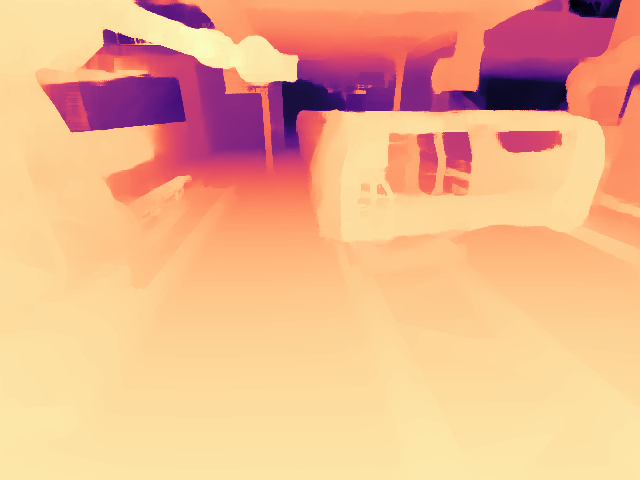}};
        \draw[red,line width=0.05cm, dashed] (0.02, 0.4) rectangle ++(2.60, 2.60);
    \end{tikzpicture}
    &
    \begin{tikzpicture}
        \node[anchor=south west,inner sep=0] at (0,0) {\includegraphics[width=0.33\linewidth]{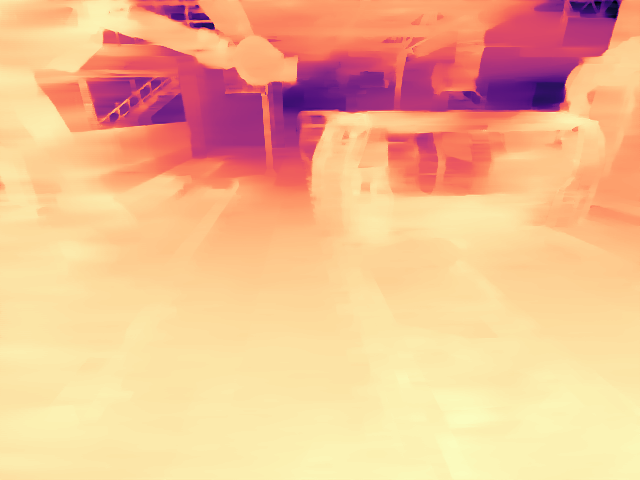}};
        \draw[red,line width=0.05cm, dashed] (0.02, 0.4) rectangle ++(2.60, 2.60);
    \end{tikzpicture}
    &
    \begin{tikzpicture}
        \node[anchor=south west,inner sep=0] at (0,0) {\includegraphics[width=0.33\linewidth]{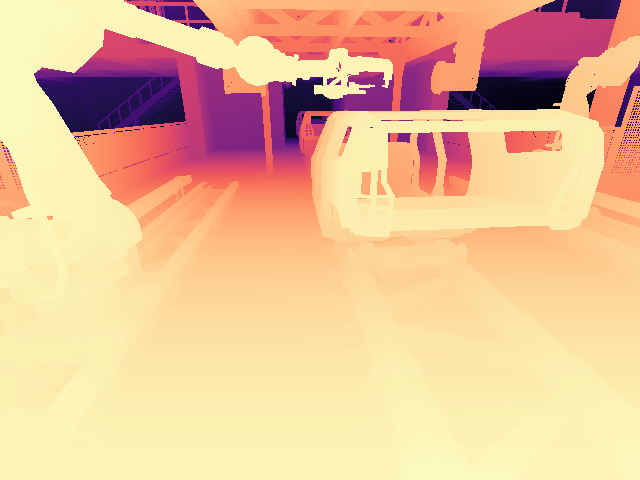}};
        \draw[red,line width=0.05cm, dashed] (0.02, 0.4) rectangle ++(2.60, 2.60);
    \end{tikzpicture}   \\
    \tiny \textbf{Target} ($\scriptstyle t$) & \tiny \textbf{Source} ($\scriptstyle t-N$) & \tiny \textbf{Sparse Depth} (${\scriptscriptstyle t-N \rightarrow t}$)  
    \\
    \begin{tikzpicture}
        \node[anchor=south west,inner sep=0] at (0,0) {\includegraphics[width=0.33\linewidth]{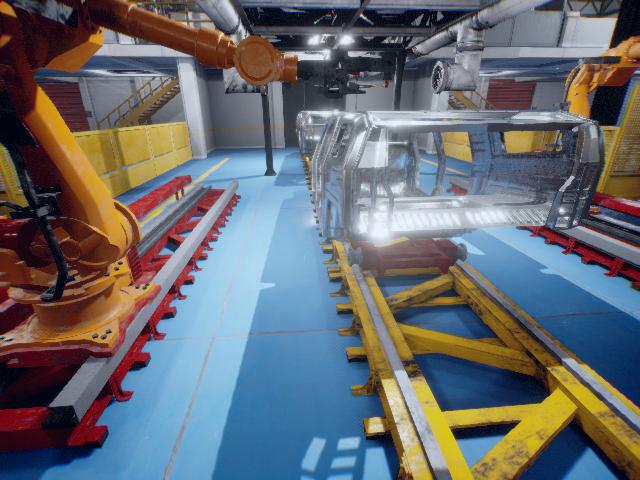}};
        \draw[red,line width=0.05cm, dashed] (0.02, 0.4) rectangle ++(2.64, 2.64);
    \end{tikzpicture} 
    & 
    \begin{tikzpicture}
        \node[anchor=south west,inner sep=0] at (0,0) {\includegraphics[width=0.33\linewidth]{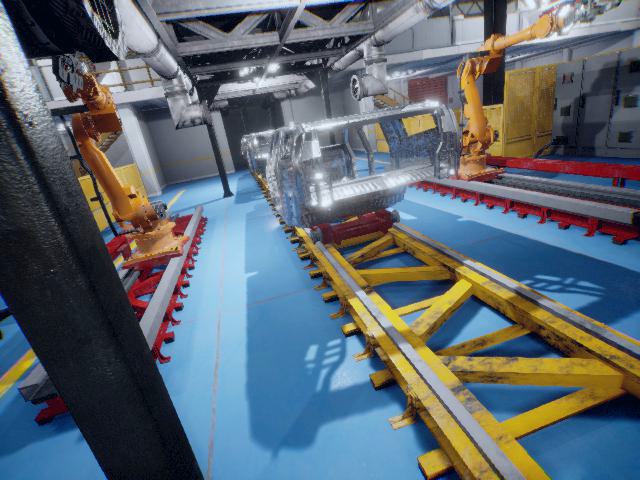}};
        \draw[red,line width=0.05cm, dashed] (0.3, 1.1) rectangle ++(1.0, 1.5);
        \begin{scope}[overlay]
        \draw [red,line width=0.05cm, dashed, ->] (0.18, 1.35) -- (-1.2, 1.15);
        \end{scope}
    \end{tikzpicture}
    &
    \includegraphics[width=0.33\linewidth]{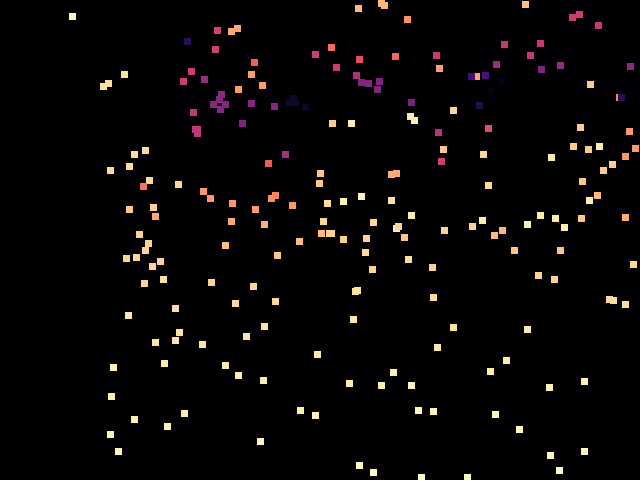} \\
    \multicolumn{3}{c}{\tiny \textbf{3D Point Cloud}} \\
    \multicolumn{3}{c}{
    \includegraphics[trim=3cm 5.5cm 5cm 9cm, clip, width=\linewidth]{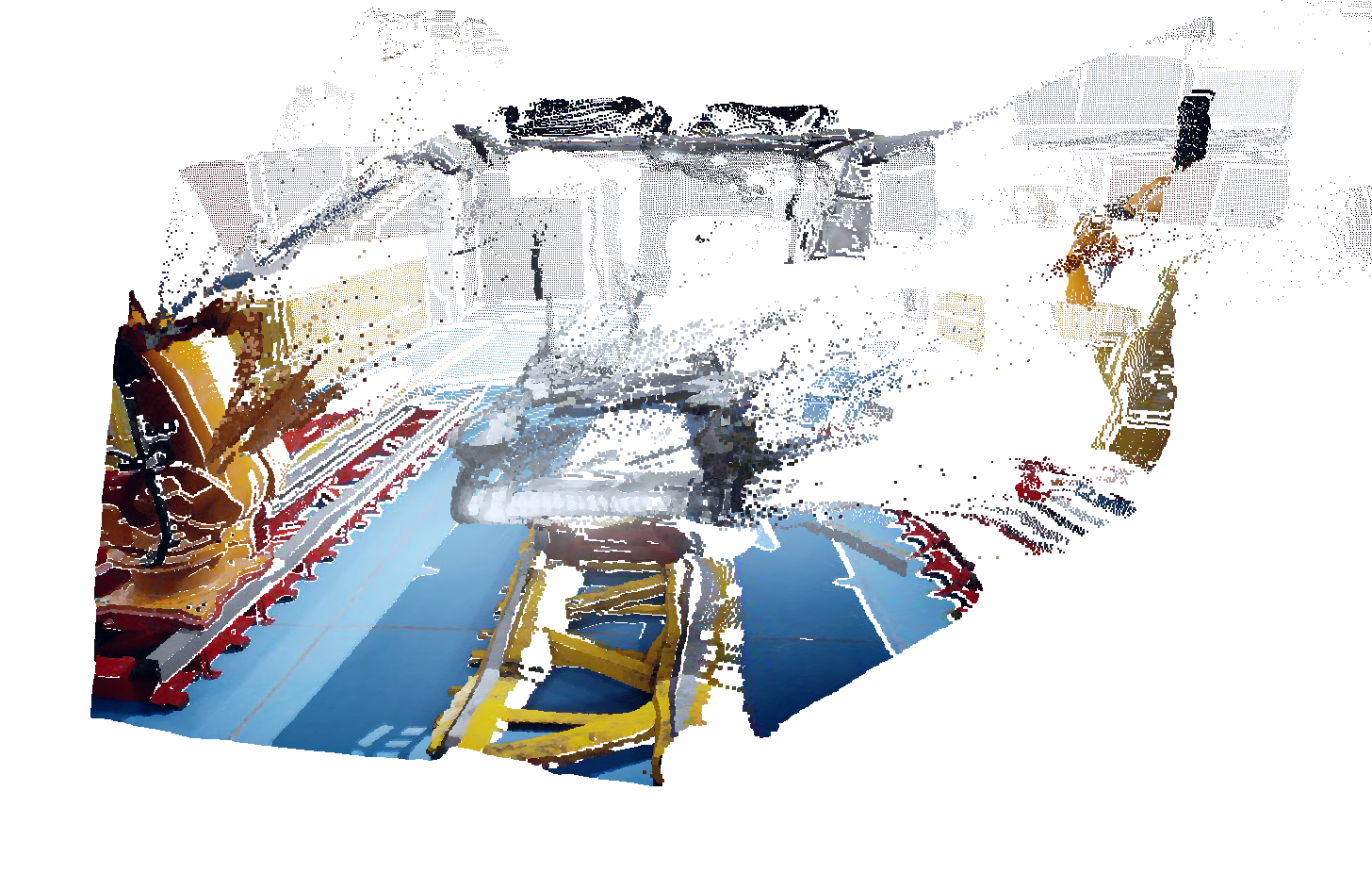}
    } \\
    \end{tabular}\vspace{-0.3cm}
    \caption{\textbf{Moving Objects.} Example of our framework behavior on moving objects in a video sequence from TartanAir \cite{tartanair2020iros}. On top, ours and \cite{nlspn} depth estimation. Below, are the target, source, and depth points used. In the dashed red bounding box is highlighted a robotic arm moving regardless of the camera. Last, is the 3D projection of our depth estimation. Our approach provides consistent monocular depth estimation where multi-view and depth cues are wrong.}\vspace{-0.2cm}
    \label{fig:moving-objects}
    \vspace{-0.5cm}
\end{figure}

\subsection{Pose Noise Sensitivity}
We propose an additional experiment against test-time pose noise sensitivity, in the event that the pose estimator -- \eg, any visual-inertial odometry or SLAM pipeline -- may introduce an error in the pose estimate. Let us represent a pose by a six degrees of freedom vector $\mathbf{q} := (\mathbf{t}, \mathbf{r}) \in \mathbb{R}^6$ with translation $\mathbf{t}$ and rotation $\mathbf{r}$ (in Euler angles). Given each ground truth pose $\mathbf{q}$, we draw random Gaussian noise on it, yielding the perturbed pose $\hat{\mathbf{q}} \sim \mathcal{N}(\mathbf{q}, \lambda^2 \operatorname{diag}({\mathbf{q}})^2)$ where we dub $\lambda$ the pose noise factor. We feed $\hat{\mathbf{q}}$ everywhere we would use $\mathbf{q}$ in the pipeline. At the top of Figure~\ref{fig:pose-noise-study}, we report a visual example where we project a known sparse depth point with a set of 100 noisy poses drawn from the aforementioned distribution with pose noise factor $\lambda = 0.3$ for a given $\mathbf{q}$.
Such noise not only affects our pipeline but any other depth completion method as well, in the assumption that the pose is used to reproject the sparse depth points since it corresponds to a significant uncertainty in the depth hints localization. In our case, the impact of pose errors is more subtle; first off, it affects the geometry cues, in that the epipolar correlation block samples along perturbed epipolar lines. Secondly, it affects the reprojected sparse depth points on the target view, as per the completion case.
At the bottom of Figure \ref{fig:pose-noise-study} we report quantitative results sweeping $\lambda \in [0, 0.3]$ (where $0$ is the noiseless case). Each model in this evaluation is trained with noise-free pose on TartanAir \cite{tartanair2020iros}. As it would seem that we could be more affected by pose errors, by this test we assess that the gap in performance between our method and depth completion methods remains fixed w.r.t. $\lambda$, \ie, in a fair evaluation pose noise causes a degradation that increases gracefully with $\lambda$ whilst keeping an almost fixed quality gap between all methods.

\subsection{Moving Objects}

Furthermore, we qualitatively assess the behavior of our approach on scenes with moving objects. Traditionally, multi-view methods work under the assumption of a static environment, to enable the use of multi-view geometry cues. Nonetheless, moving objects can occur in real use-cases. In the automotive scenario, other vehicles move \cite{Geiger2012CVPR}; in the indoor scenario people or objects can move. In this paper, we do not focus on the challenge of dealing with scene motion. Nonetheless, we acknowledge the existence of this issue and thus we provide a qualitative study of how our framework behaves in moving areas. Figure \ref{fig:moving-objects} provides an example scene from TartanAir \cite{tartanair2020iros} where a robotic arm moves on an assembly line. When sparse depth points are projected from the source view $I_{t-N}$ to the target view $I_{t}$ sparse depth points gathered on the arm are wrongly projected. Moreover, multi-view cues are not useful in this case -- \ie even if the network predicts the correct depth on the target view for a moving object the projection on the source view leads to a wrong position. Thus, the monocular features are the unique useful information to estimate depth in the dashed red box. Our approach demonstrates to better exploit such information than NLSPN \cite{nlspn}, effectively ignoring misleading multi-view and sparse depth information. Nonetheless, monocular depth estimated from a non-specialized approach is far from being fully accurate, as can be observed in the point cloud at the bottom of Figure \ref{fig:moving-objects}.

\section{Qualitative Results}

\subsection{3D Reconstruction}

\begin{figure*}[h]
    \centering
    \begin{tabular}{@{}c@{}c@{}c@{}}
    \includegraphics[width=0.30\linewidth]{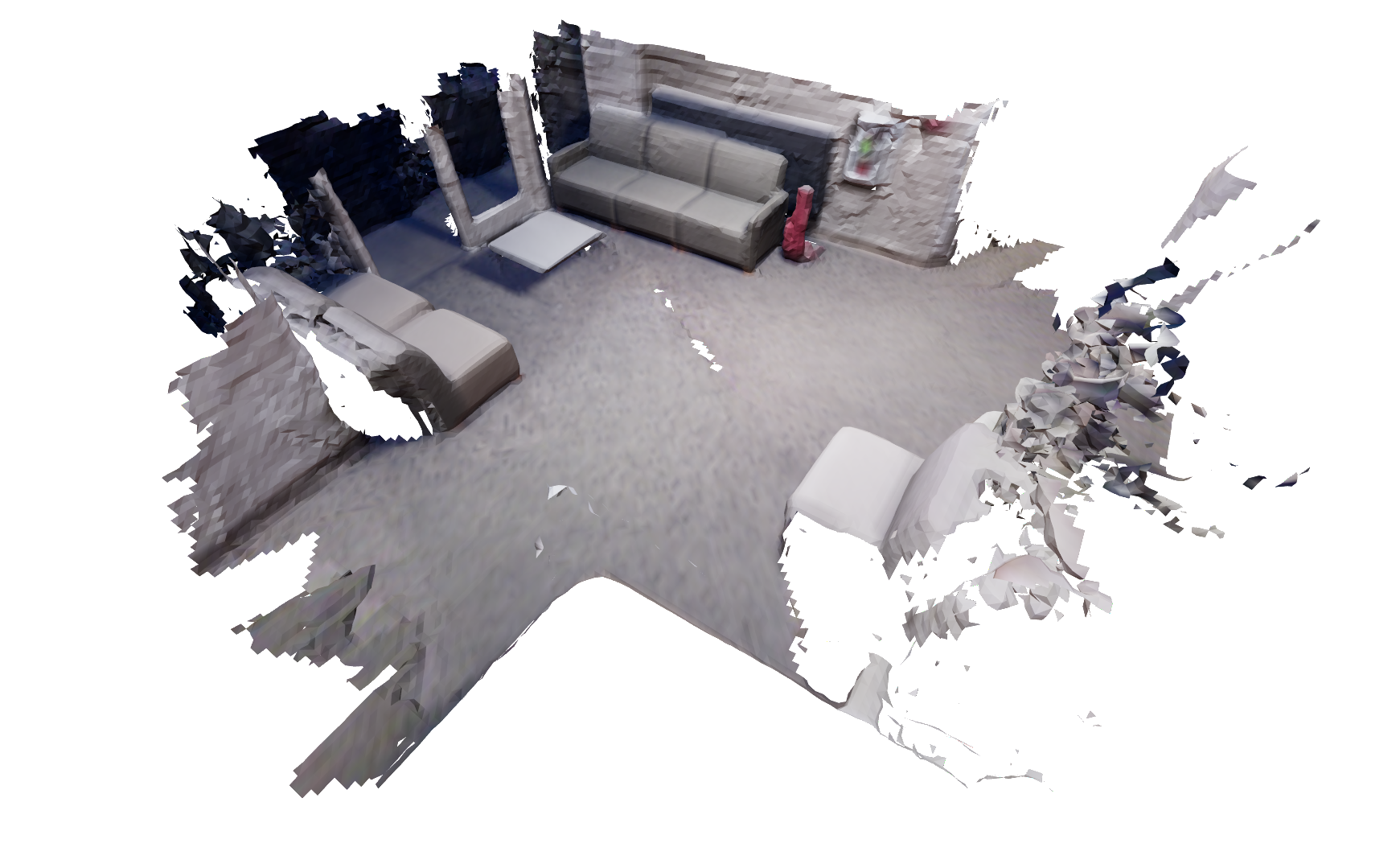} &
    \includegraphics[width=0.30\linewidth]{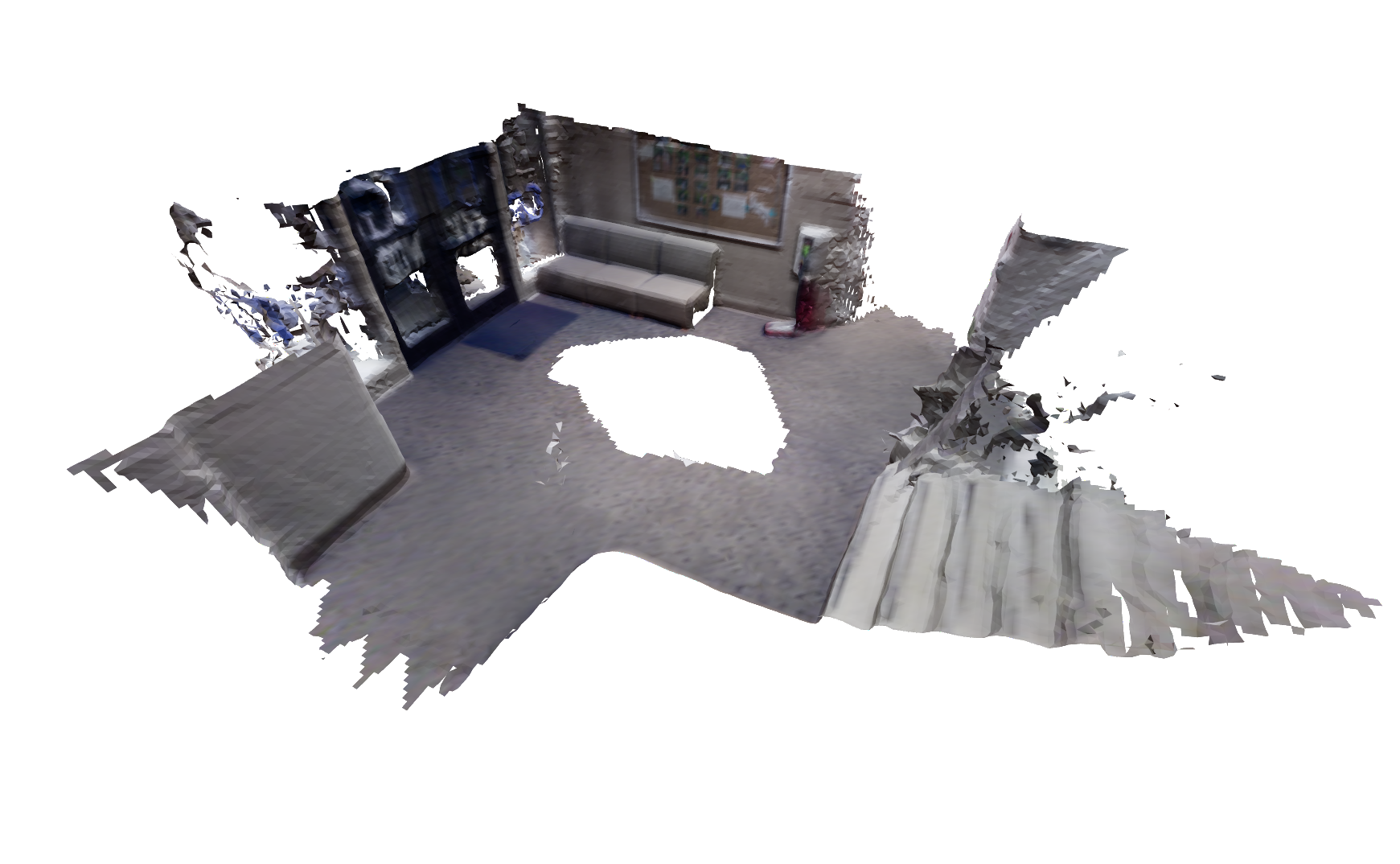} &
    \includegraphics[width=0.30\linewidth]{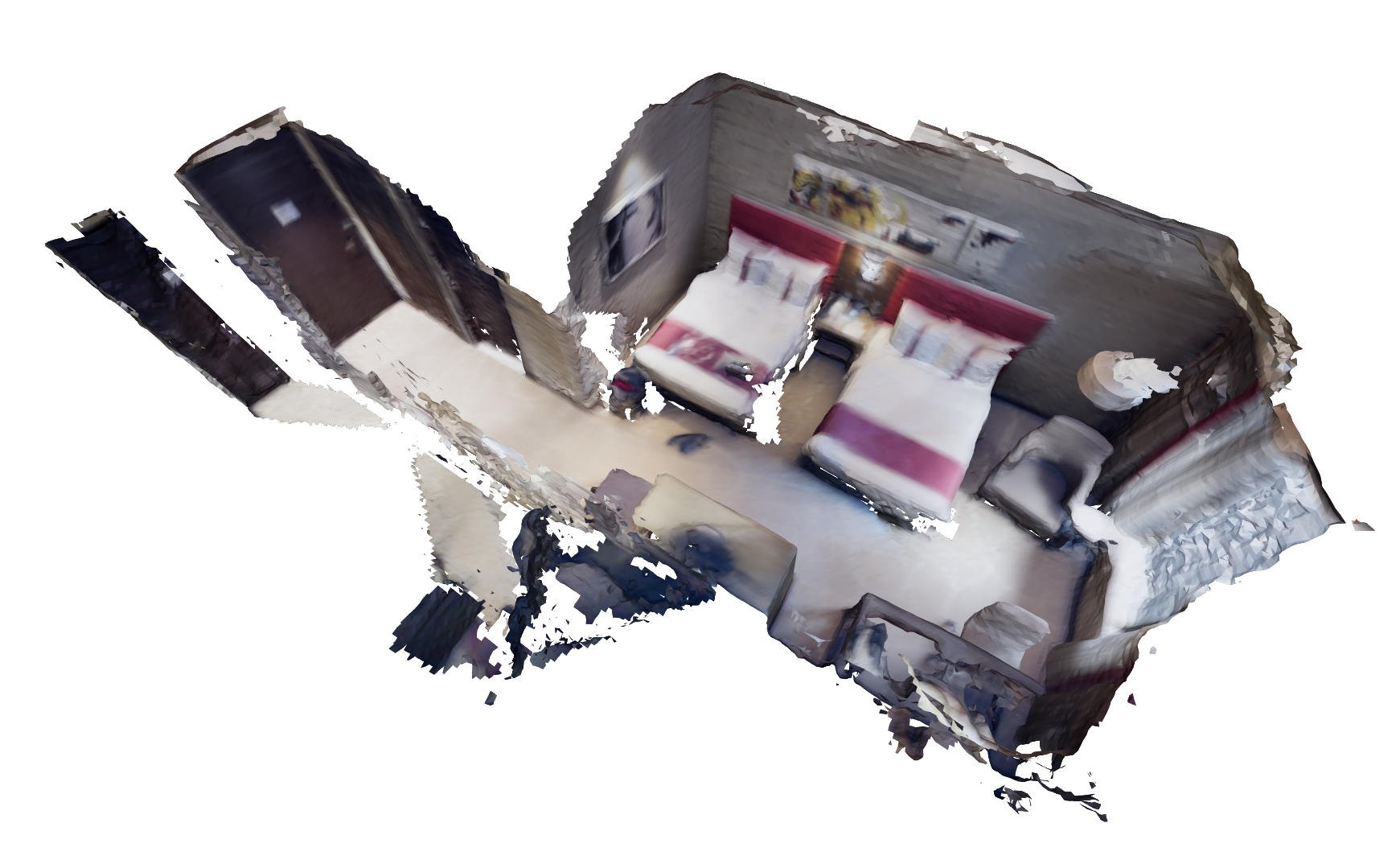} \\
    \tiny \textbf{Scene 0714}  & \tiny \textbf {Scene 0715} & \tiny \textbf{Scene 0721}         \\
    \includegraphics[width=0.30\linewidth]{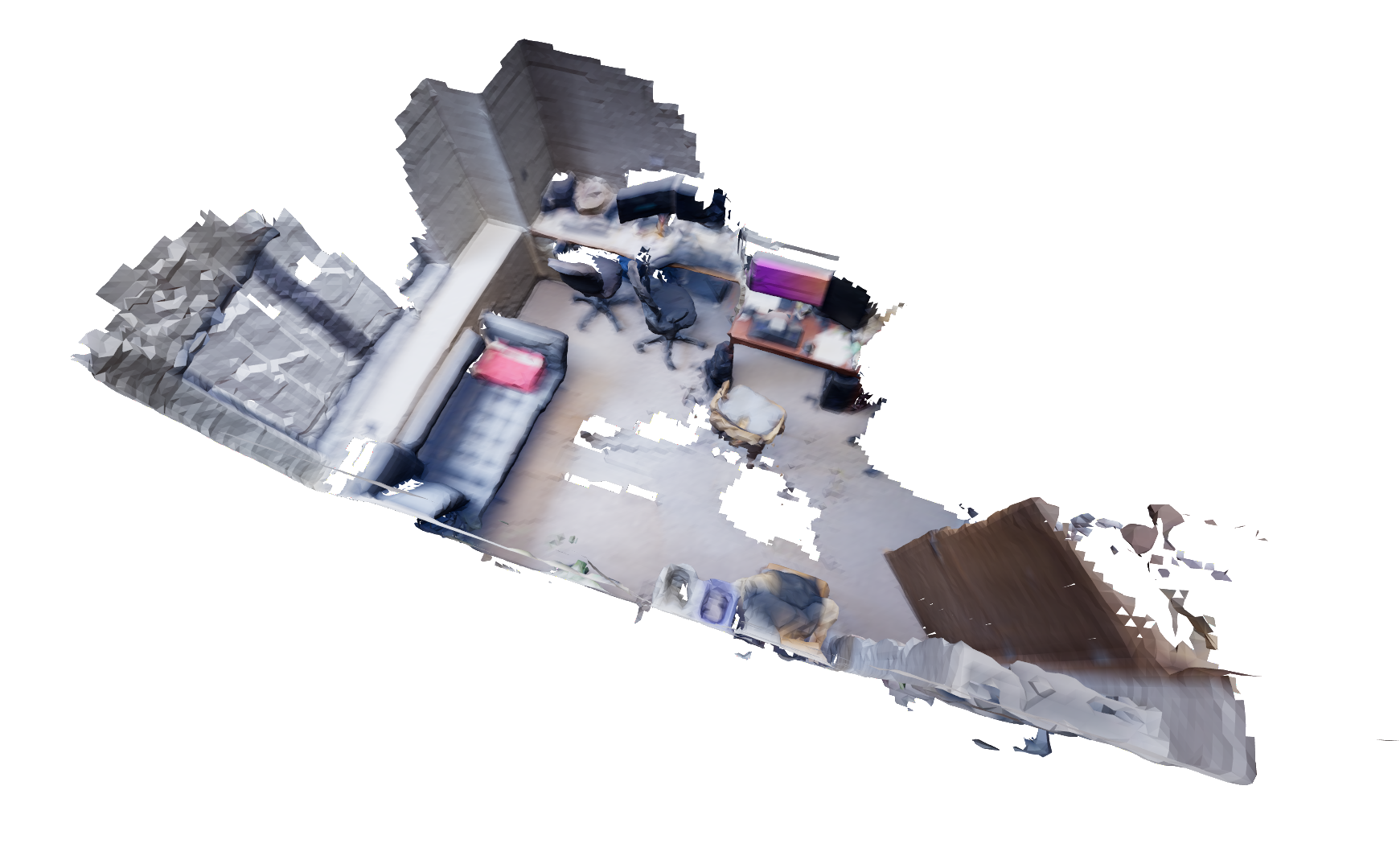} &
    \includegraphics[width=0.30\linewidth]{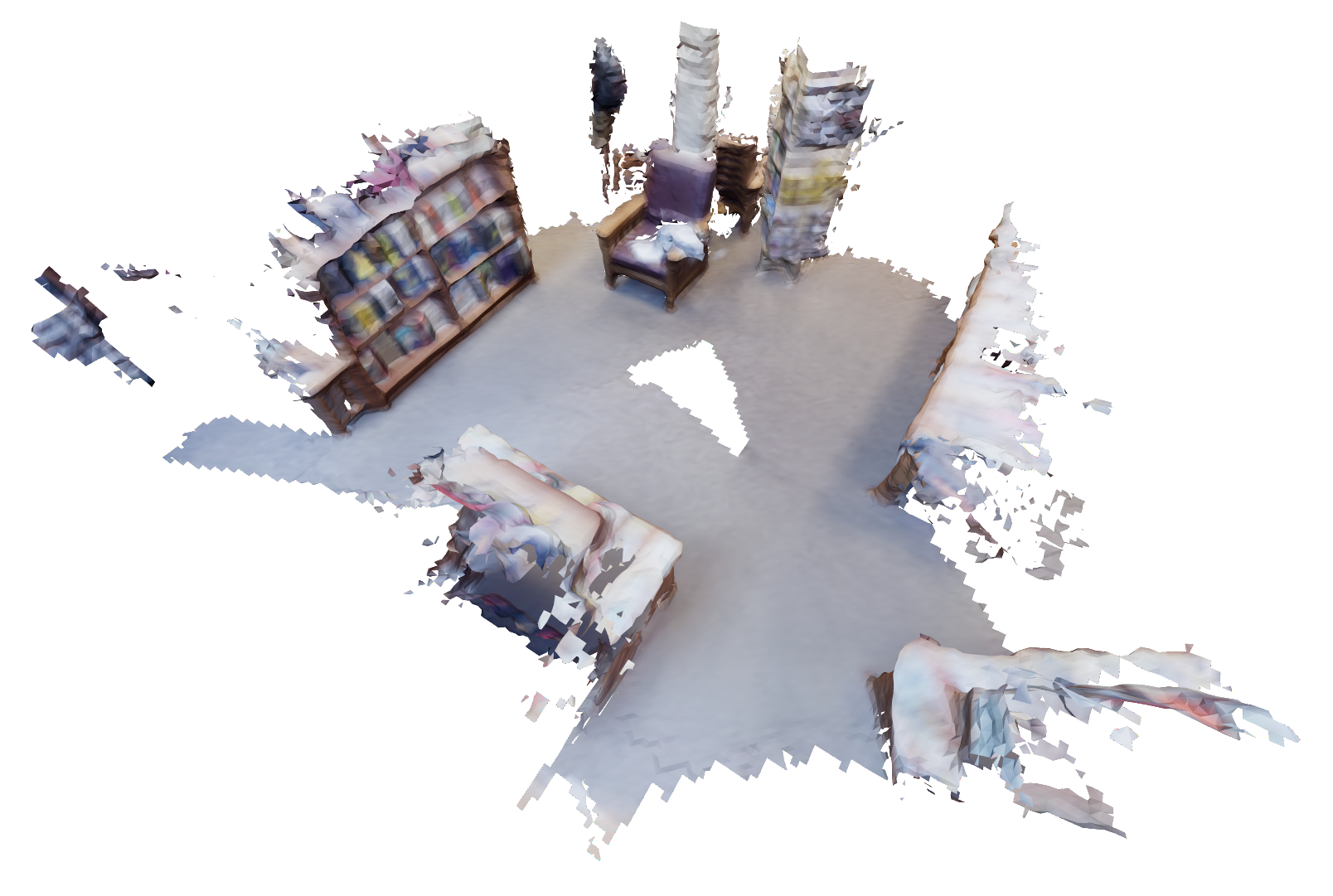} &
    \includegraphics[width=0.30\linewidth]{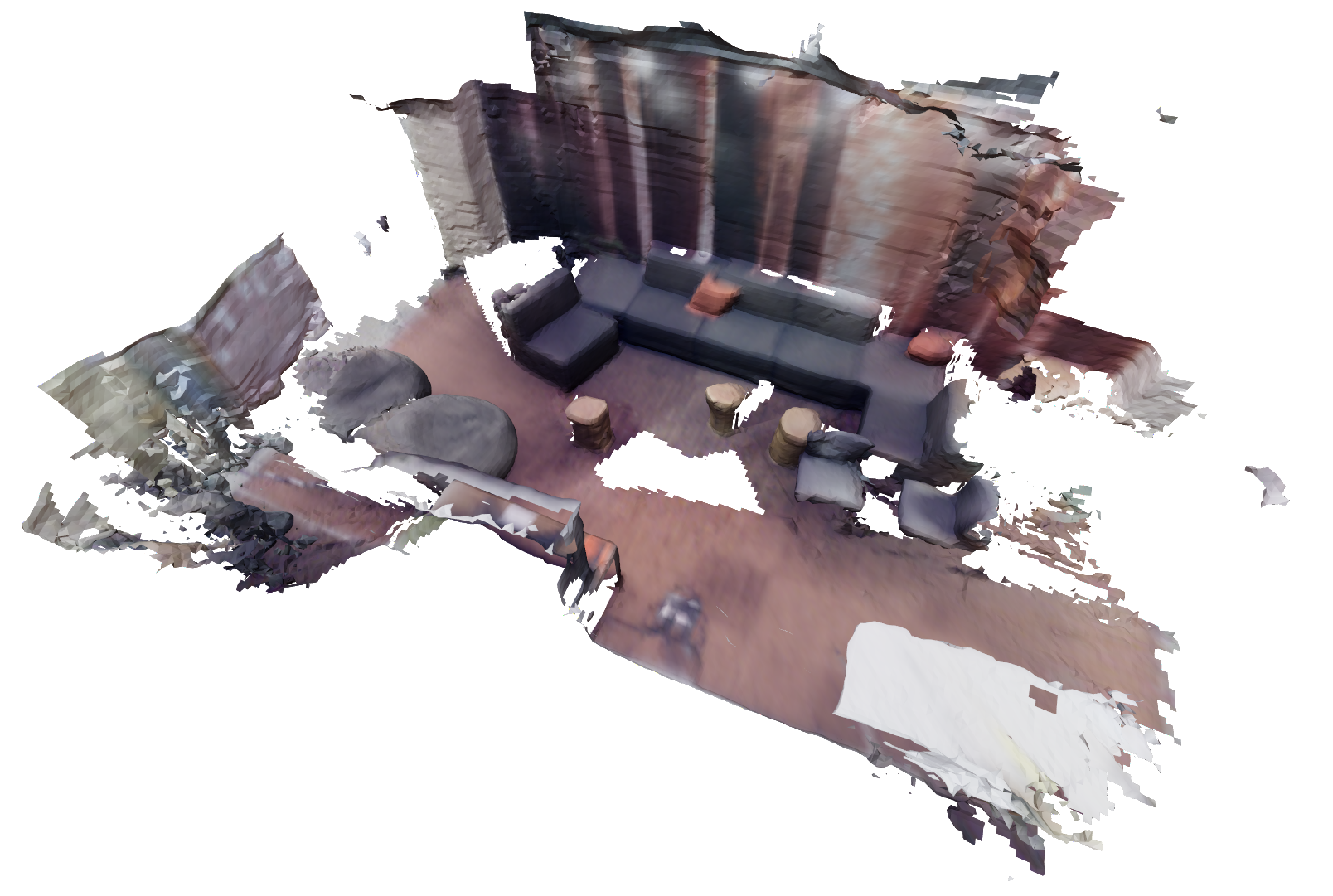} \\
    \tiny \textbf{Scene 0745}  & \tiny \textbf {Scene 0800} & \tiny \textbf{Scene 0805}         \\
    \end{tabular}
    \caption{\textbf{ScanNetV2 3D Reconstruction.} We provide different qualitative mesh reconstructions on ScanNetV2 \cite{dai2017ScanNet} in different indoor scenarios. Our approach enables fine-grain effective RGB-D 3D reconstruction exploiting both high frame rate RGB cameras and slow yet accurate sparse active sensors.}
    \label{fig:scannet-qualitatives}
\end{figure*}

We provide qualitative results about the final 3D reconstructions we obtain through our approach in indoor environments from ScanNetV2 \cite{dai2017ScanNet} and 7Scenes \cite{Shotton2013SceneCR}. To build each mesh we use 500 sparse depth points and sparsification ratio $\tau = 0.2$. Then, we integrate depth prediction in a TSDF volume using the same parameters as \cite{sayed2022simplerecon} and extract the mesh with the marching cubes algorithm. Qualitatives are showed in Figure \ref{fig:scannet-qualitatives} and Figure \ref{fig:sevenscenes-qualitatives} for ScanNetV2 and 7Scenes, respectively.

\begin{figure*}[h]
    \centering
    \begin{tabular}{@{}c@{}c@{}c@{}}
    \includegraphics[width=0.30\linewidth]{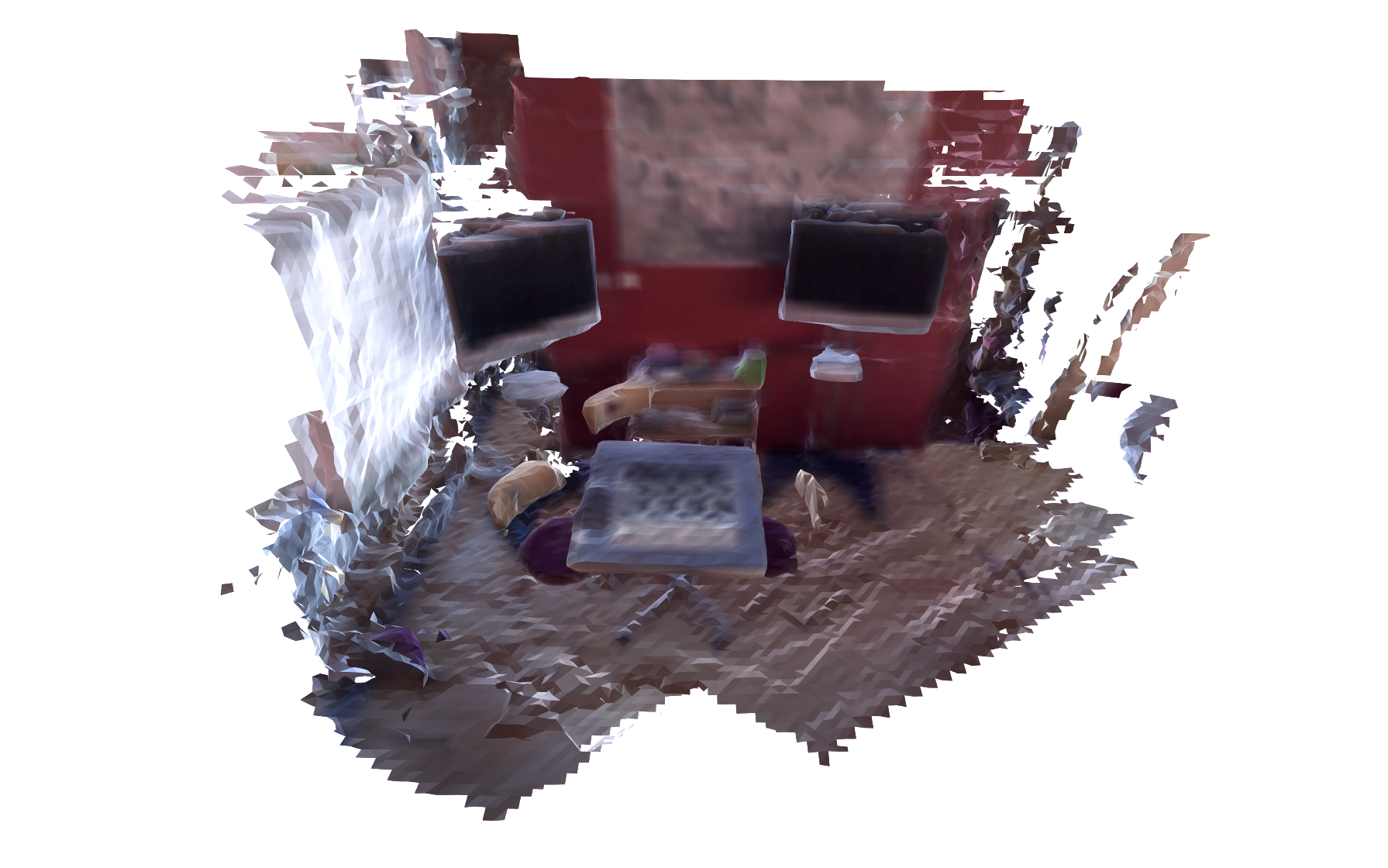}      &
    \includegraphics[width=0.30\linewidth]{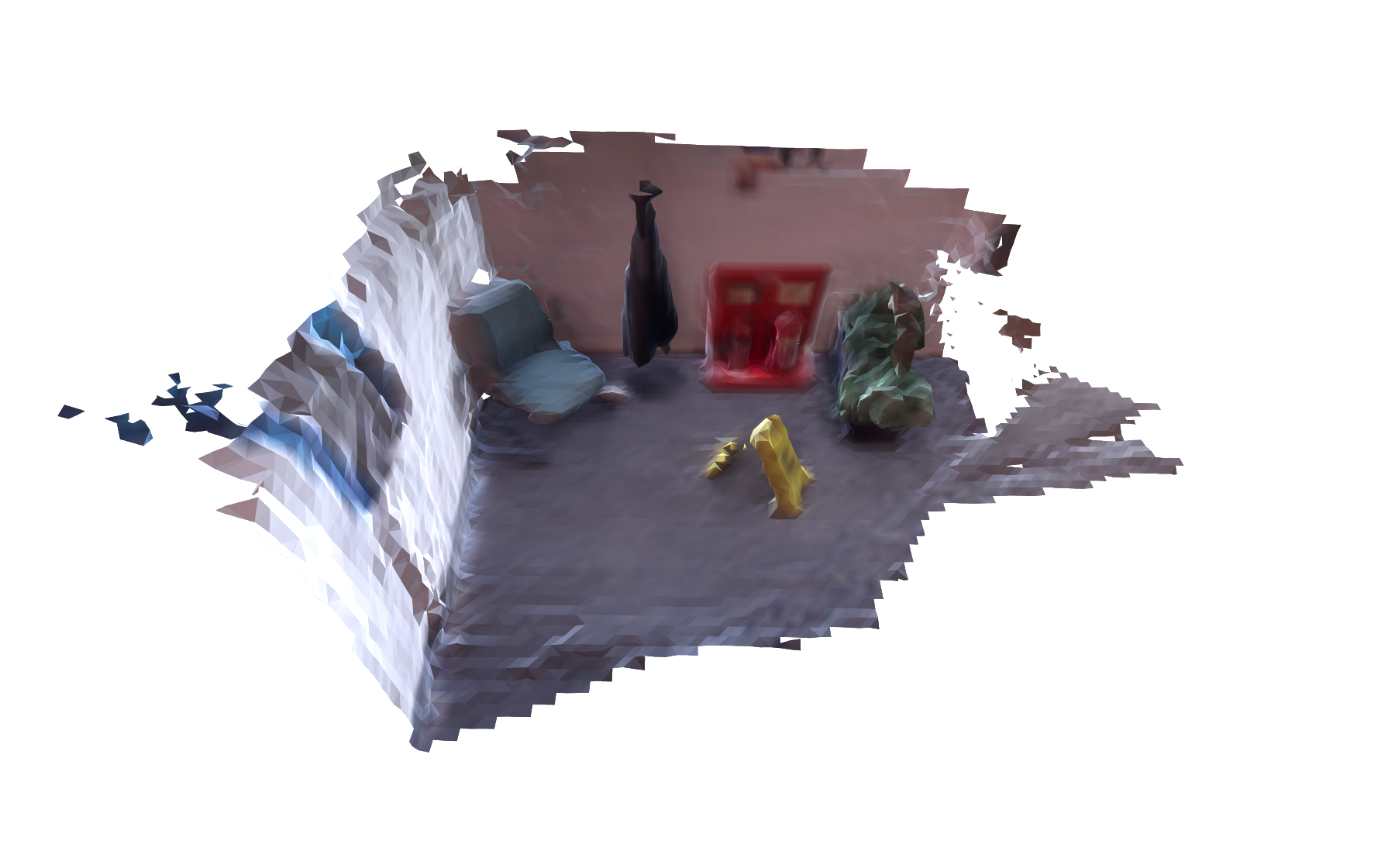}       &
    \includegraphics[width=0.30\linewidth]{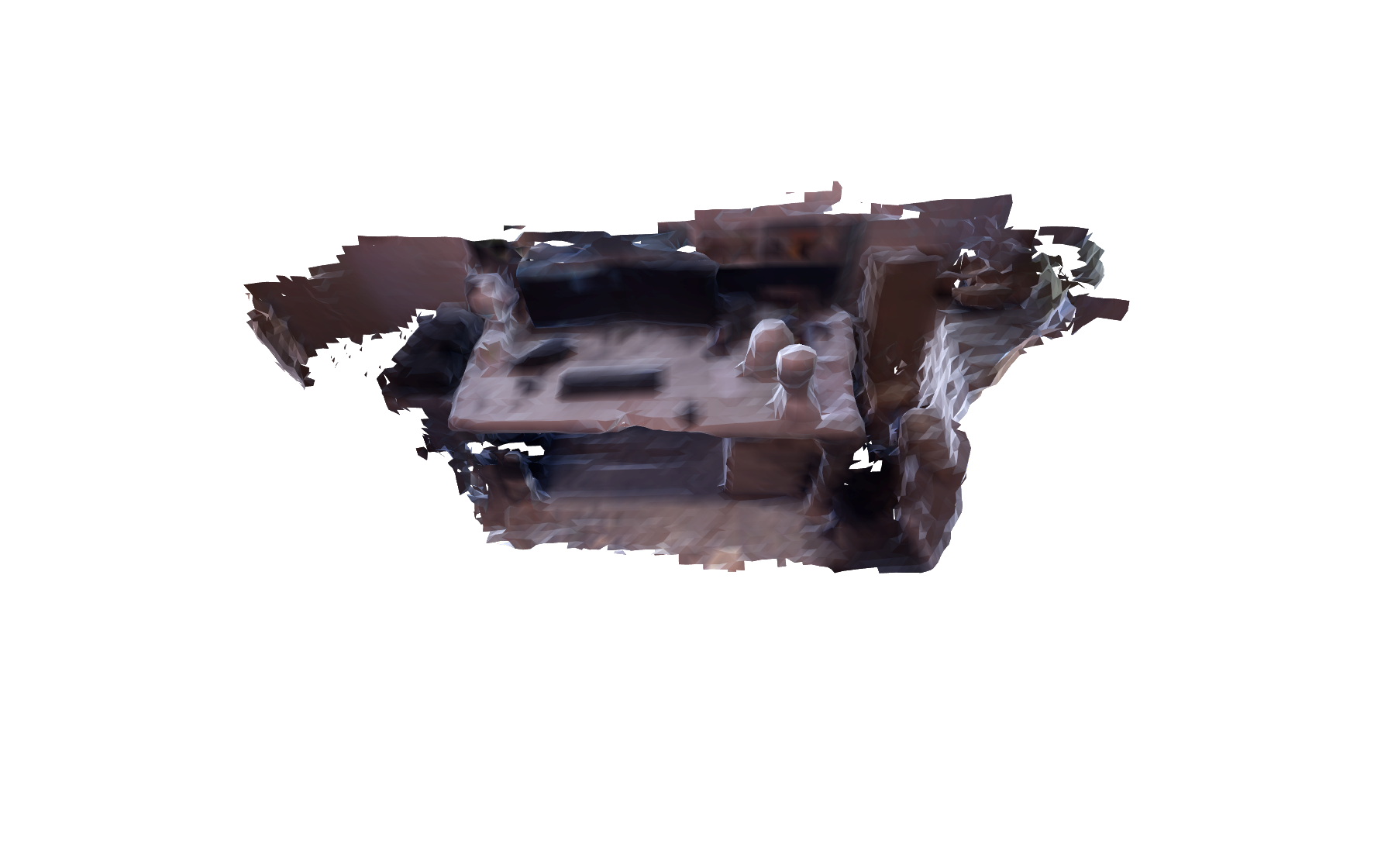}      \\
    \tiny \textbf{Chess}  & \tiny \textbf {Fire} & \tiny \textbf{Heads}                                \\
    \includegraphics[width=0.30\linewidth]{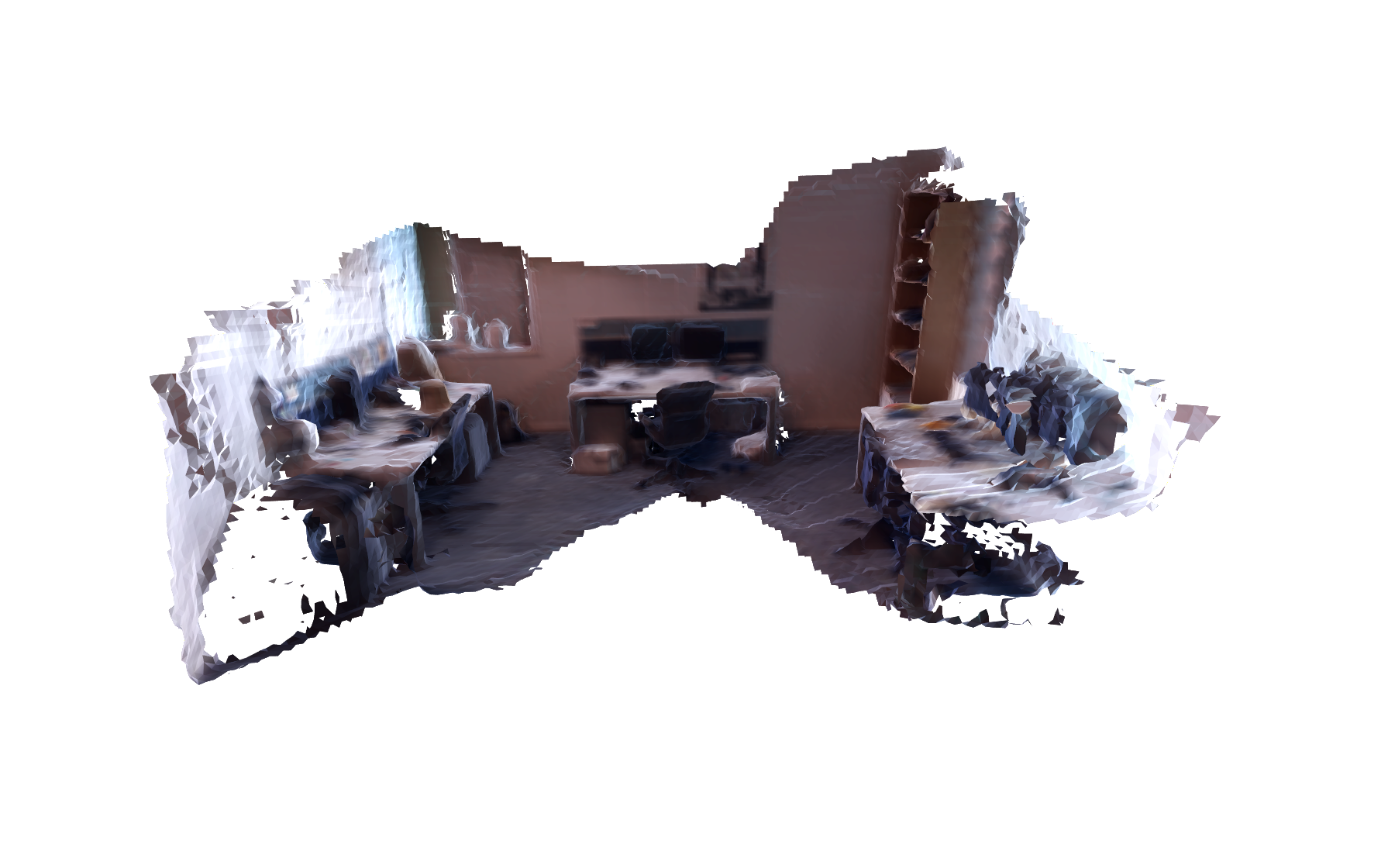}     &
    \includegraphics[width=0.30\linewidth]{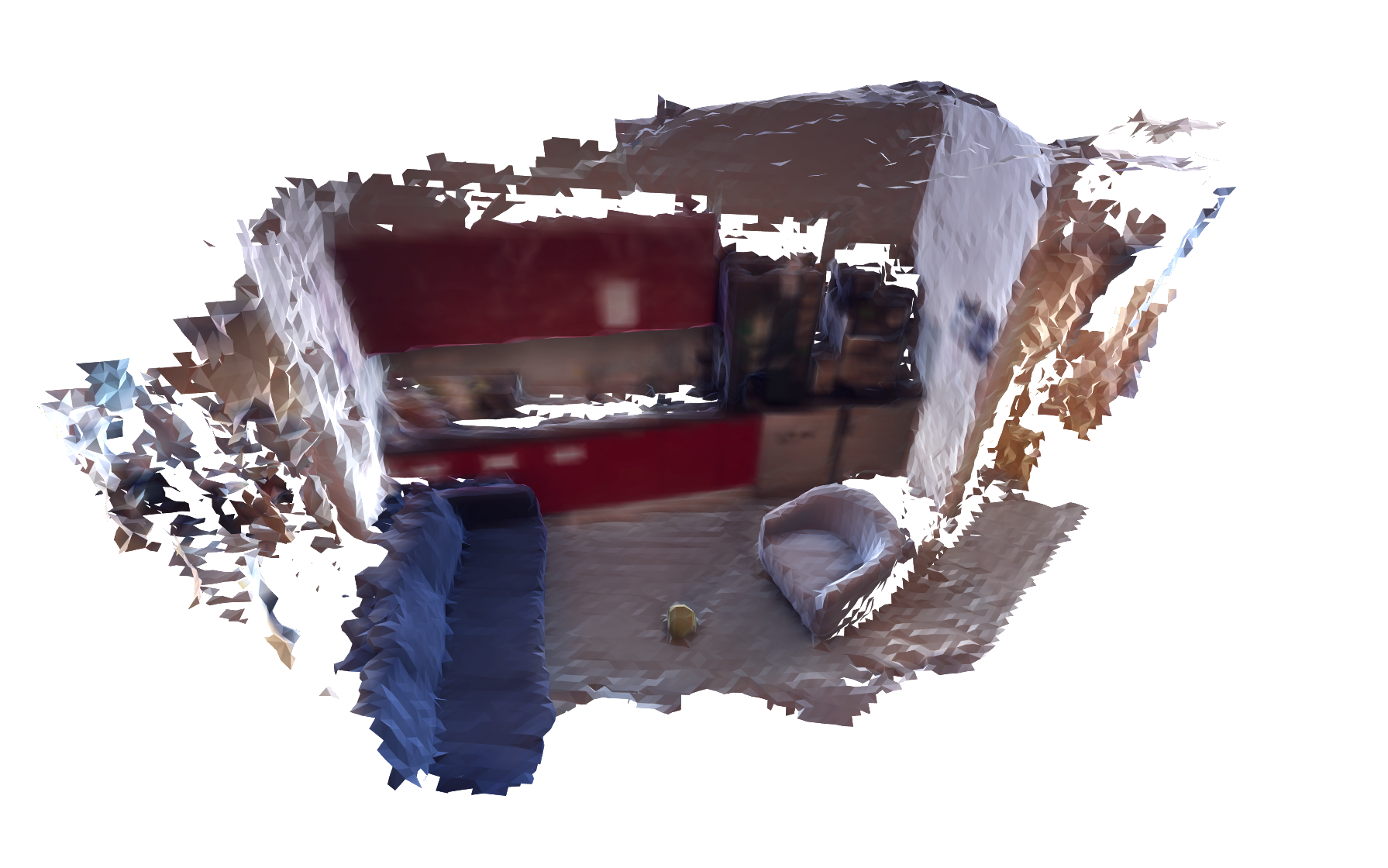}    &
    \includegraphics[width=0.30\linewidth]{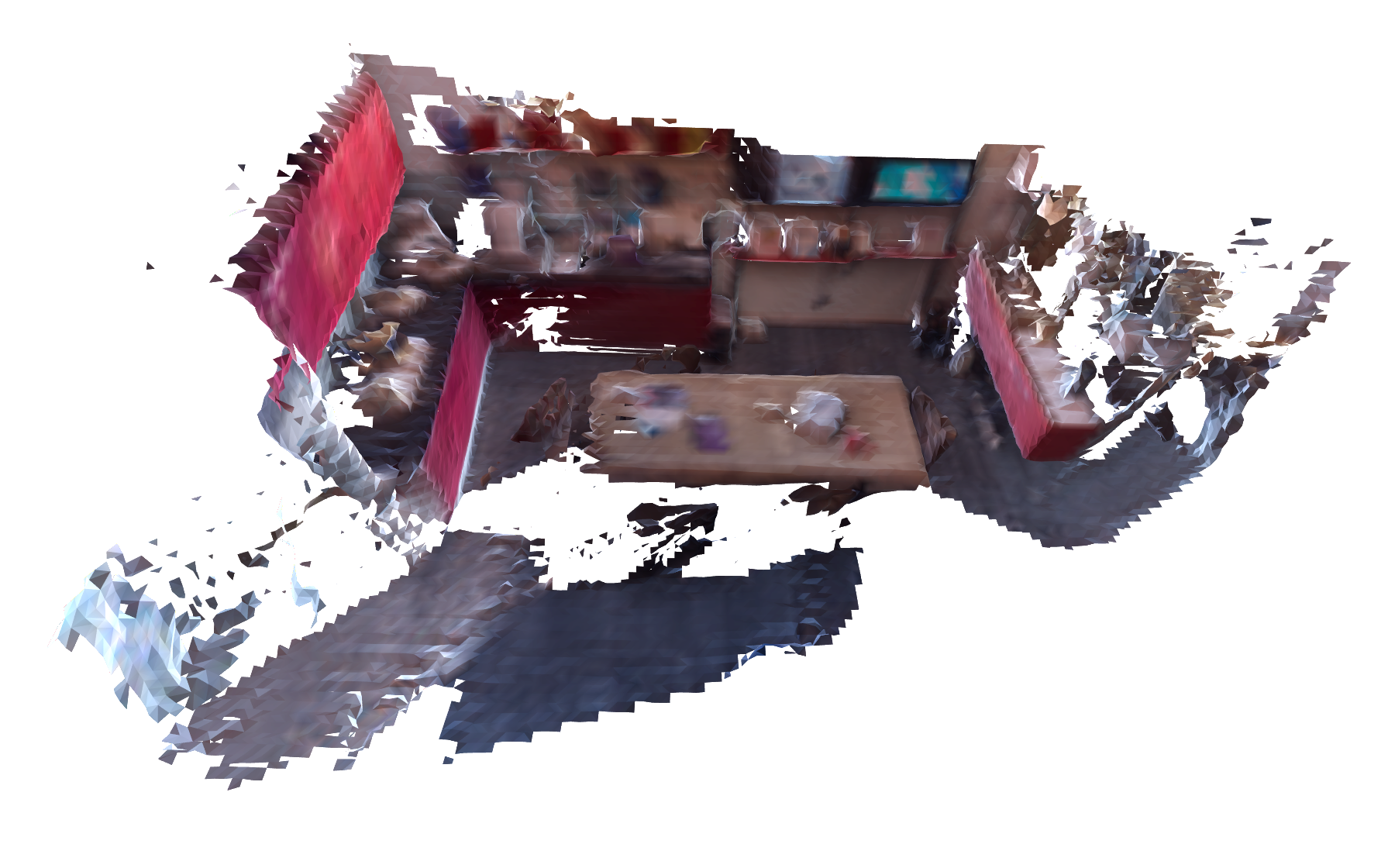} \\
    \tiny \textbf{Office}  & \tiny \textbf {Pumpkin} & \tiny \textbf{Red Kitchen}                      \\
    \end{tabular}
    \caption{\textbf{7Scenes 3D Reconstruction.} We provide qualitative mesh reconstructions on 7Scenes \cite{Shotton2013SceneCR} in generalization -- \ie we train only on ScanNetV2 \cite{dai2017ScanNet}. Our framework demonstrates the capability to easily generalize to new environments as assessed in the experiments on 7Scenes.}
    \label{fig:sevenscenes-qualitatives}
\end{figure*}

\subsection{Generalization on Waymo}

\begin{figure}[h!]
    \centering
    \setlength{\tabcolsep}{1pt}
    \begin{tabular}{@{}ccc@{}}
    \tiny{\textbf Source View} & \tiny{\textbf Target View} & \tiny{\textbf Prediction} \\
    \includegraphics[width=0.33\linewidth]{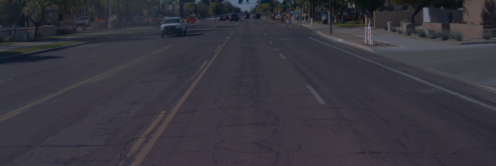}
    & \includegraphics[width=0.33\linewidth]{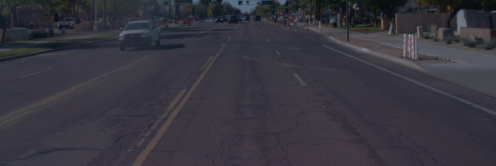}
    & \includegraphics[width=0.33\linewidth]{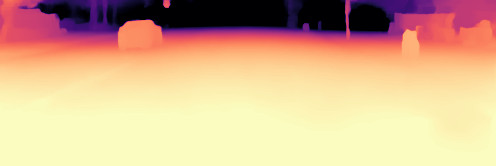}\vspace{-0.1cm} \\
    \includegraphics[width=0.33\linewidth]{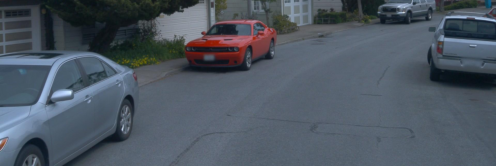}
    & \includegraphics[width=0.33\linewidth]{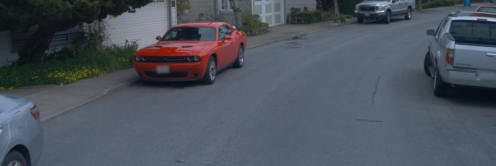}
    & \includegraphics[width=0.33\linewidth]{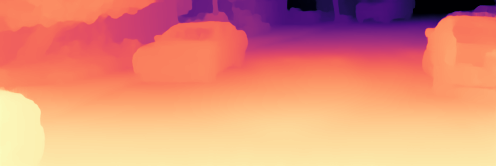}\vspace{-0.1cm} \\
    \includegraphics[width=0.33\linewidth]{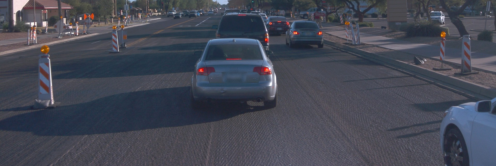}
    & \includegraphics[width=0.33\linewidth]{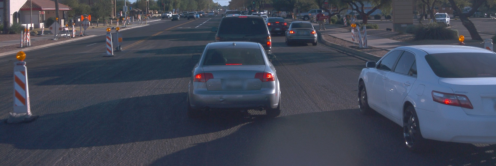}
    & \includegraphics[width=0.33\linewidth]{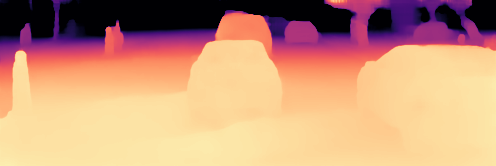}
    \end{tabular}
    \caption{\textbf{Qualitatives on Waymo.} We test DoD trained on KITTI on the Waymo dataset in generalization. Although the KITTI's small size and repetitiveness hamper deployment in generalization DoD produces reasonable results predictions}
    \label{fig:waymo-qualitatives}
\end{figure}

Finally, in Figure \ref{fig:waymo-qualitatives} we provide a few inferences of DoD trained on KITTI in generalization on the Waymo Dataset. DoD provides reasonable reconstructions even in this scenario, despite the small size and repetitiveness of the KITTI dataset hamper good generalization.

\end{document}